\documentclass[11pt]{article}
\usepackage[letterpaper,margin=1in]{geometry}
\usepackage[T1]{fontenc}
\usepackage[utf8]{inputenc}
\usepackage{lmodern}
\usepackage[protrusion=true,expansion=false]{microtype}
\usepackage{amsmath,amssymb}
\usepackage{graphicx}
\usepackage{booktabs}
\usepackage{multirow}
\usepackage{adjustbox}
\usepackage{subcaption}
\usepackage{booktabs}
\usepackage{array}
\usepackage{xspace}
\usepackage{tikz}
\usetikzlibrary{arrows.meta,positioning,fit,calc}
\usepackage{pgfplots}
\pgfplotsset{compat=1.18}
\usepackage[pagebackref=false,breaklinks,colorlinks,allcolors=blue]{hyperref}
\hypersetup{pdftitle={LG-FiLM: Local-Global Mixture-of-Experts Modulation for RAW-to-RGB Reconstruction},pdfauthor={Fei Li, Xilei Cao, Shichang Zhai, Kun Zhan},pdfsubject={},pdfkeywords={}}
\usepackage{xcolor}
\definecolor{plotblue}{rgb}{0.21,0.49,0.74}
\definecolor{plotgreen}{rgb}{0.13,0.48,0.40}
\newcommand{\method}{LG-FiLM\xspace}
\newcommand{\raw}{\mathbf{x}}

\newcommand{\pred}{\hat{\mathbf{y}}}
\newcommand{\feat}{\mathbf{F}}
\newcommand{\ctx}{\mathbf{z}}

\newcommand{\clip}{\operatorname{clip}}

\begin{document}

\title{\textbf{LG-FiLM: Local--Global Mixture-of-Experts\\ Modulation for RAW-to-RGB Reconstruction}}

\author{Fei Li \quad Xilei Cao \quad Shichang Zhai \quad Kun Zhan\\
LiAuto, China\\
\texttt{\{lifei3,caoxilei,zhaishichang,zhankun\}@lixiang.com}}

\date{}
\maketitle

\begin{abstract}
An image signal processor does more than reconstruct an image. It also has to decide how strongly to denoise, how to balance color, and which tone curve maps the scene onto the display, and these settings depend on the scene itself: the curve that suits a backlit sky is not the one that suits an indoor portrait. Most neural ISPs learn a single fixed transformation and apply it to every scene. We instead make the adaptation an explicit part of the network. Our model, \method, keeps one shared encoder--decoder and places a small mixture of affine experts at its bottleneck; the experts' scales, shifts, and mixing weights are predicted from local neighborhood evidence together with a global scene descriptor. Reconstruction stays in a single shared pathway and only the processing mode is switched. We evaluate \method{} on three datasets: Fujifilm UltraISP, Zurich RAW to RGB, and a COCO-derived benchmark built with an inverse ISP. The model reaches the best reconstruction fidelity of the compared entries on the primary benchmark and transfers to the other two.
\end{abstract}

\section{Introduction}
\label{sec:intro}

The image signal processor of a modern smartphone camera is responsible for turning a raw Bayer measurement into the photograph the user sees. It demosaics, denoises, restores color, and maps the scene onto a displayable tonal range, and each of these decisions has to be made per capture. This is the central difficulty of the problem: there is no single ISP setting that is right for every scene. Denoising strength has to follow the noise level, color gains have to follow the illuminant, and tone mapping has to reconcile two requirements that pull in opposite directions---global compression must accommodate the luminance range of the whole scene, while local contrast must preserve detail inside regions of very different brightness~\cite{reinhard2002photographic,durand2002bilateral}. A backlit sky and an indoor portrait need different curves, and a single fixed configuration cannot serve both.

Current neural ISPs address this by training one network end to end on paired RAW--RGB data~\cite{schwartz2019deepisp,chen2018seeing,ignatov2020pynet,li2024rmfa}. The network learns from the data, but it applies the same weights to every input: whatever scene dependence appears is a by-product of fitting the training distribution, not a mechanism the architecture provides. Conditioning and attention make the dependence more visible~\cite{souza2023metaisp,ignatov2022pynetv2}, but they usually inject a single image-level signal, which cannot describe regions that need different corrections, while purely local evidence has the opposite limitation and cannot determine the appearance the scene is meant to have. What is missing is an explicit mechanism that selects among alternative processing behaviors according to the scene.

We propose LG-FiLM, a model that introduces a mixture-of-experts strategy at the bottleneck of a shared encoder--decoder to fill exactly this gap. The experts are small affine transformations of the bottleneck feature, and a router predicts, for every location, how strongly each expert should contribute; both the router and the experts are driven jointly by a global scene descriptor and by local neighborhood evidence. All experts are evaluated and their outputs are fused with soft weights, so the composed transformation moves smoothly between processing modes instead of switching between discrete branches, and the encoder, decoder and skip paths remain shared and unduplicated. Adaptation therefore becomes an explicit, low-dimensional object in the network, one that can be inspected and ablated, at a cost of 0.268\,M parameters on top of a 1.402\,M backbone.

We evaluate the approach on three datasets that probe different aspects of the problem. Fujifilm UltraISP, which pairs mobile-sensor RAW with Fujifilm RGB references, is the primary benchmark and the setting for our comparison with reported methods. Zurich RAW to RGB evaluates reconstruction from Huawei P20 RAW measurements to Canon 5D Mark IV RGB photographs. A COCO-derived benchmark, built by applying an inverse ISP to ordinary photographs, checks that the design carries over to a structurally different RAW domain. Across all three, LG-FiLM produces high-fidelity reconstructions.

We contribute (i) a formulation of input-adaptive ISP processing as a dense mixture of affine feature transformations inside a shared reconstruction network, which separates conditional modulation from reconstruction; (ii) a local--global conditioning design that drives both the expert transformations and their mixing weights at 0.268\,M additional parameters, with no duplicated encoder or decoder; and (iii) an experimental analysis of reconstruction quality, the roles of expert transformations and routing, and inference cost across three benchmarks.

\begin{figure}[t]
\centering
\definecolor{lgink}{HTML}{26354A}
\definecolor{lgpanel}{HTML}{F8FAFC}
\definecolor{lgborder}{HTML}{CBD5E1}
\definecolor{lgblue}{HTML}{427BB8}
\definecolor{lgteal}{HTML}{258C84}
\definecolor{lgviolet}{HTML}{8062AB}
\definecolor{lgorange}{HTML}{CE8630}
\definecolor{lgrose}{HTML}{BD627E}
\resizebox{\linewidth}{!}{%
\begin{tikzpicture}[
  font=\sffamily\fontsize{8}{9.5}\selectfont,text=lgink,
  block/.style={rounded corners=2pt,line width=0.65pt,align=center,
    minimum height=0.78cm,inner sep=3pt},
  encoder/.style={block,draw=lgblue,fill=lgblue!12,text width=1.20cm},
  decoder/.style={block,draw=lgteal,fill=lgteal!12,text width=1.20cm},
  flow/.style={-{Latex[length=1.6mm,width=1.1mm]},draw=lgink,line width=0.75pt},
  skip/.style={flow,draw=lgteal,dashed,rounded corners=2pt},
  weight/.style={flow,draw=lgviolet,dashed,rounded corners=2pt},
  title/.style={font=\sffamily\bfseries\fontsize{10}{11}\selectfont,anchor=west},
  note/.style={font=\sffamily\fontsize{7}{8}\selectfont,align=center},
  port/.style={circle,draw=lgteal,fill=white,inner sep=0pt,
    minimum size=0.23cm,font=\sffamily\fontsize{8}{8}\selectfont}]
\draw[draw=lgborder,fill=lgpanel,rounded corners=4pt,line width=0.6pt]
  (0,4.18) rectangle (19,7.95);
\draw[draw=lgborder,fill=lgpanel,rounded corners=4pt,line width=0.6pt]
  (0,0) rectangle (19,3.86);
\node[title] at (0.30,7.66) {(a) Network architecture};
\node[anchor=east,font=\sffamily\bfseries\fontsize{8}{9}\selectfont,
  text=lgorange] at (18.68,7.66) {Default configuration: 8 experts};
\node[note,text=lgblue] at (3.8,7.33) {SHARED ENCODER};
\node[note,text=lgteal] at (15.1,7.33) {SHARED DECODER};
\node[note,text=lgteal] at (9.5,7.33) {$\oplus$ Additive skip connections};
\foreach \i in {0,...,3} {
  \foreach \j in {0,...,3} {
    \pgfmathtruncatemacro{\bayercell}{mod(\i,2)+2*mod(\j,2)}
    \ifcase\bayercell\def\cellcolor{lgteal}\or\def\cellcolor{lgblue}%
      \or\def\cellcolor{lgrose}\or\def\cellcolor{lgteal}\fi
    \fill[\cellcolor!65,draw=white,line width=0.3pt]
      (0.28+\i*0.19,6.20+\j*0.19) rectangle ++(0.19,0.19);
  }
}
\node[note,font=\sffamily\fontsize{6.3}{8}\selectfont] at (0.66,5.90)
  {Sparse RAW\\$3,\;H$};
\coordinate (rawout) at (1.06,6.58);
\node[encoder,text width=1.08cm] (stem) at (2.05,6.58)
  {\textbf{Stem}\\$16,\;H$};
\node[encoder] (enc1) at (3.90,6.32) {\textbf{Enc. 1}\\$32,\;H/2$};
\node[encoder] (enc2) at (5.50,5.78) {\textbf{Enc. 2}\\$64,\;H/4$};
\node[encoder] (enc3) at (7.10,5.24) {\textbf{Enc. 3}\\$128,\;H/8$};
\node[block,draw=lgorange,fill=lgorange!17,line width=1pt,
  text width=2.30cm,minimum height=0.92cm] (bottleneck) at (9.50,5.24)
  {\textbf{LG-FiLM}\\$128\!\rightarrow\!256,\;H/8$\\[-1pt]
   {\fontsize{7}{8}\selectfont 8 affine experts}};
\node[decoder] (dec3) at (11.90,5.24) {\textbf{Dec. 3}\\$128,\;H/8$};
\node[decoder] (dec2) at (13.50,5.78) {\textbf{Dec. 2}\\$64,\;H/4$};
\node[decoder] (dec1) at (15.10,6.32) {\textbf{Dec. 1}\\$32,\;H/2$};
\node[block,draw=lgrose,fill=lgrose!12,text width=1.12cm] (head) at (16.90,6.58)
  {\textbf{Head}\\PS$\times2$};
\fill[lgblue!65,draw=white] (17.99,6.29) rectangle (18.55,6.89);
\fill[lgteal!65,draw=white] (18.10,6.20) rectangle (18.66,6.80);
\fill[lgrose!65,draw=white] (18.21,6.11) rectangle (18.77,6.71);
\node[note] at (18.38,5.90) {RGB\\$3,\;H$};
\draw[flow] (rawout)--(stem.west);
\foreach \a/\b in {stem/enc1,enc1/enc2,enc2/enc3,enc3/bottleneck,
  bottleneck/dec3,dec3/dec2,dec2/dec1,dec1/head} {
  \draw[flow] (\a.east)--(\b.west);
}
\draw[flow] (head.east)--(17.97,6.58);
\foreach \n/\height in {1/7.02,2/6.48,3/5.94} {
  \node[port] (add\n) at (dec\n.center |- 0,\height) {$+$};
  \draw[skip] (enc\n.north)--(enc\n.center |- 0,\height)
    --(add\n.west);
  \draw[draw=lgteal,line width=0.75pt] (add\n.south)--(dec\n.north);
}
\node[note,anchor=west] at (0.32,4.47)
  {Encoder: stride-2 Conv + 2 RBs\quad Decoder: Conv + resize + $\oplus$ + RB};
\node[note,anchor=east] at (18.68,4.47)
  {RB: residual block\quad Head: Conv + PS$\times2$ + ReLU};
\draw[draw=lgorange,dashed,line width=0.8pt]
  (bottleneck.south)--(9.50,4.64);
\draw[flow,draw=lgorange,dashed] (9.50,4.16)--(9.50,3.88);
\node[title] at (0.30,3.57) {(b) LG-FiLM bottleneck};
\node[note,anchor=east] at (18.68,3.57)
  {$H=W=256:\quad h=w=32,\quad C=128,\quad N=1{,}024$ tokens};
\node[block,draw=lgblue,fill=lgblue!9,text width=0.73cm,
  minimum height=0.68cm] (feature) at (0.62,1.94) {$\feat_{\rm enc}$\\$128$};
\node[encoder,text width=1.08cm,minimum height=0.68cm] (shared) at (2.04,1.94)
  {Shared\\RBs $\times2$};
\coordinate (f) at (3.03,1.94);
\fill[lgink] (f) circle (1.2pt);
\node[above right=1pt] at (f) {$\feat$};
\node[block,draw=lgteal,fill=lgteal!12,text width=1.48cm] (local) at (4.50,2.62)
  {\textbf{Local}\\{\fontsize{7}{8}\selectfont $3\!\times\!3$ DWConv}};
\node[block,draw=lgviolet,fill=lgviolet!12,text width=1.48cm] (global) at (4.50,1.18)
  {\textbf{Global}\\{\fontsize{6.7}{8}\selectfont GAP + broadcast}};
\node[block,draw=lgviolet,fill=lgviolet!7,text width=1.48cm] (concat) at (6.64,1.94)
  {\textbf{Concat}\\$\ctx_i=[\mathbf u_i;\mathbf g]$};
\node[block,draw=lgviolet,fill=lgviolet!15,text width=2.15cm] (router) at (9.02,2.68)
  {\textbf{Router}\\{\fontsize{7}{8}\selectfont Linear + softmax}\\$p_{i,1},\ldots,p_{i,8}$};
\node[block,draw=lgorange,fill=lgorange!6,text width=2.62cm,
  minimum height=1.55cm,inner sep=0pt] (experts) at (9.02,1.13) {};
\node[note,font=\sffamily\bfseries\fontsize{8}{9}\selectfont] at (9.02,1.65)
  {8 affine experts};
\foreach \k/\x/\y in {1/8.10/1.27,2/8.71/1.27,3/9.32/1.27,4/9.93/1.27,
                        5/8.10/0.86,6/8.71/0.86,7/9.32/0.86,8/9.93/0.86} {
  \node[draw=lgorange!75,fill=lgorange!18,rounded corners=1pt,
    minimum width=0.47cm,minimum height=0.30cm,inner sep=1pt,note] at (\x,\y) {$e_{\k}$};
}
\node[note] at (9.02,0.51) {MLP $\rightarrow(\gamma_{i,k},\beta_{i,k})$};
\node[block,draw=lgorange,fill=lgorange!13,text width=2.22cm,
  minimum height=1.02cm] (fusion) at (12.02,1.94)
  {\textbf{Soft fusion}\\
   $\bar\gamma_i=\sum_k p_{i,k}\gamma_{i,k}$\\
   $\bar\beta_i=\sum_k p_{i,k}\beta_{i,k}$};
\node[block,draw=lgrose,fill=lgrose!12,text width=2.24cm,
  minimum height=1.02cm] (affine) at (15.02,1.94)
  {\textbf{Modulation}\\
   $\feat_i'=\feat_i\odot(1+\bar\gamma_i)$\\$+\;\bar\beta_i$};
\node[block,draw=lgblue,fill=lgblue!12,text width=1.25cm,
  font=\sffamily\fontsize{7}{8}\selectfont] (projection) at (17.37,1.94)
  {$1\!\times\!1$ Conv\\$128\!\rightarrow\!256$};
\draw[flow] (feature)--(shared);
\draw[draw=lgink,line width=0.75pt] (shared.east)--(f);
\draw[flow] (f)|-(local.west);
\draw[flow] (f)|-(global.west);
\draw[flow,draw=lgteal] (local.east)-|(concat.north);
\draw[flow,draw=lgviolet] (global.east)-|(concat.south);
\coordinate (condition) at (7.63,1.94);
\draw[draw=lgink,line width=0.75pt] (concat.east)--(condition);
\draw[flow] (condition)|-(router.west);
\draw[flow] (condition)|-(experts.west);
\draw[weight] (router.east)-|(fusion.north);
\draw[flow,draw=lgorange] (experts.east)-|(fusion.south);
\draw[flow,draw=lgorange] (fusion)--(affine);
\draw[flow] (affine)--(projection);
\draw[flow] (projection.east)--(18.57,1.94);
\node[note] at (18.51,1.48) {$\feat_{\rm out}$};
\draw[flow,rounded corners=2pt] (f)--(3.03,0.19)--(15.02,0.19)--(affine.south);
\node[note,fill=lgpanel,inner sep=2pt] at (12.40,0.19) {Shared feature $\feat_i$};
\node[note,text=lgorange] at (12.02,0.91) {All experts, every token};
\node[note,text=lgviolet] at (6.64,0.92) {$\ctx_i\in\mathbb R^{2C}$};
\end{tikzpicture}}
\caption{Architecture of \method. (a) Sparse three-channel RAW preserves sensor-grid alignment through a shared encoder--decoder with three additive skips. Stage labels give channels and height; width scales identically. The deepest decoder stage preserves resolution. (b) Local and global context jointly condition softmax attention over eight affine experts and their transformations. Soft fusion mixes scale and shift parameters to modulate the shared feature before projection. Expert output heads are zero-initialized.}
\label{fig:architecture}
\end{figure}
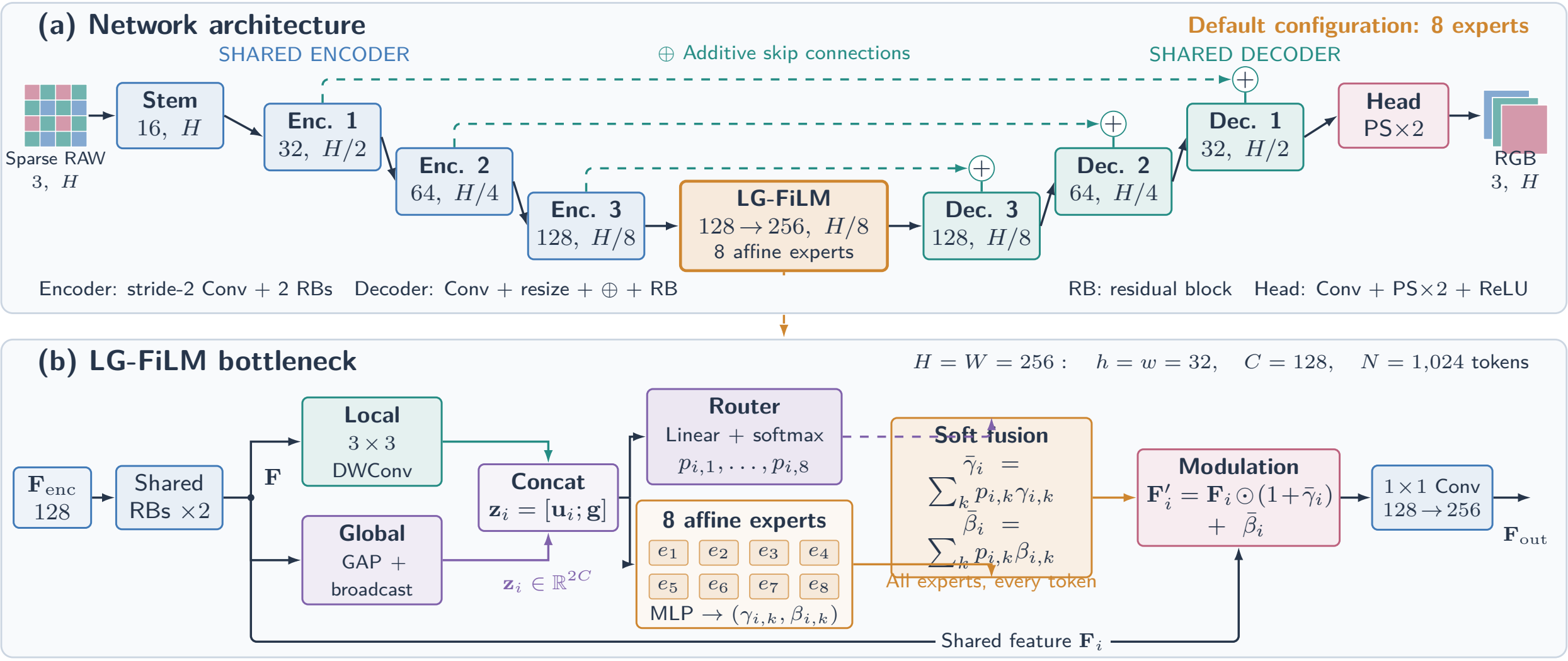

\section{Related Work}
\label{sec:related}

\paragraph{Learned image signal processing.}
DeepISP combines low-level restoration with higher-level color adjustment~\cite{schwartz2019deepisp}, and Learning to See in the Dark reconstructs long-exposure appearance from short-exposure RAW~\cite{chen2018seeing}. PyNET renders globally at low resolution and refines texture at high resolution through a pyramid~\cite{ignatov2020pynet}; PyNET-CA adds channel attention and subpixel reconstruction~\cite{kim2020pynetca}, and AWNet combines wavelet decomposition with attention and global context~\cite{dai2020awnet}. Zhang et al.\ handle cross-camera misalignment with color mapping and optical-flow-guided supervision~\cite{zhang2021aligned}. RMFA-Net preserves sensor-grid alignment by placing RAW samples on three full-resolution planes~\cite{li2024rmfa}, and MetaISP conditions color rendition on global scene structure across devices~\cite{souza2023metaisp}.

Mobile variants trade capacity for hardware-friendly operators. PyNET-V2 Mobile uses a three-scale pyramid with grouped residual blocks and channel and spatial attention~\cite{ignatov2022pynetv2}; MicroISP processes packed Bayer inputs through lightweight output-color branches and subpixel upsampling~\cite{ignatov2022microisp}. The MAI 2025 report describes reparameterized convolutions and multiplicative interactions in NVNet, diverse-branch blocks in DBF-Net, and dual-path attention in MobileIE~\cite{ignatov2025mobileisp}. Residual encoder--decoders, simplified restoration blocks, and structural reparameterization provide the underlying primitives~\cite{ronneberger2015unet,chen2022simple,ding2021repvgg}. In the reverse direction, unprocessing synthesizes RAW data by undoing image-formation steps~\cite{brooks2019unprocessing}, and InvISP learns a reversible RAW--RGB mapping~\cite{xing2021invertible}.

\paragraph{Conditional feature modulation.}
FiLM modulates features by scale and shift~\cite{perez2018film}; AdaIN transfers channel statistics~\cite{huang2017adain}; SPADE makes affine normalization spatially dependent~\cite{park2019spade}; squeeze-and-excitation recalibrates channels from globally pooled statistics~\cite{hu2018squeeze}. Deep bilateral learning predicts locally affine transforms directly in image space from local and global cues~\cite{gharbi2017bilateral}. Our module shares this conditioning vocabulary but differs in what it conditions and in what it mixes: it predicts a \emph{set} of affine transformations in bottleneck feature space and interpolates them according to a routed mixture, rather than predicting one affine transform or one modulation vector.

\paragraph{Mixtures and dynamic transformations.}
MoE combines predictors through input-dependent gates~\cite{jacobs1991adaptive}. Sparsely gated layers activate a subset of experts~\cite{shazeer2017outrageously}, Switch Transformers route one expert per token~\cite{fedus2022switch}, V-MoE extends sparse routing to visual representations~\cite{riquelme2021vision}, and Expert Choice lets experts select tokens to control load~\cite{zhou2022expert}. These designs trade total capacity against active computation, and their central difficulty is routing balance.

Dense alternatives make a different trade. Soft MoE dispatches weighted token combinations into expert slots and redistributes the outputs~\cite{puigcerver2024soft}; CondConv and dynamic convolution mix kernel banks~\cite{yang2019condconv,chen2020dynamic}; MoCE-IR assigns restoration experts with different receptive fields and computational costs~\cite{zamfir2025moce}. \method is dense in the same spirit---every expert processes every bottleneck token---but its experts transform \emph{features} through affine scale and shift instead of replacing convolution kernels, and no token dispatch or sparse-inference acceleration is assumed.

The distinction that matters for our argument is what a mixture buys. Existing MoE designs mostly use routing to \emph{allocate computation}: a token is sent to a subset of experts, and the gain comes from spending capacity where it is needed. Here routing does not allocate anything, because every expert runs regardless. What routing supplies is a continuous \emph{interpolation} between complete processing modes, and what the experts supply is the mode itself. This is why our analysis concentrates on separating the two rather than on measuring expert utilization, and why we report measured cost.

\section{Method}
\label{sec:method}

\subsection{Formulation}
\label{sec:formulation}

Write the ISP as a conditional operator $\mathcal{T}_{\theta}(\cdot;z)$ acting on a feature field, where $z$ summarizes the capture condition. A conventional network absorbs $z$ into $\theta$ and learns one function for all conditions; a conditioning approach instead lets $z$ select among operators. Both extremes are costly, for different reasons. Making $\theta$ a function of $z$ means a hypernetwork over the entire reconstruction stack, whose output size grows with the network. Making the selection hard means the operator set must be partitioned finely enough that a single choice suffices, which forces either many large branches or a coarse, discontinuous adaptation.

We instead keep the reconstruction stack fixed and make the \emph{adaptation} a low-dimensional perturbation of a shared feature. Concretely, we ask the network to predict, at the bottleneck, a set of affine maps $\{(\boldsymbol\gamma_k,\boldsymbol\beta_k)\}_{k=1}^{E}$ along with their mixture weights, and to apply their convex combination. Two properties follow immediately and are what make the design practical. First, the perturbation has $2E$ channel vectors, a size independent of the spatial resolution and of the encoder--decoder depth, so capacity is spent on adaptation rather than on duplicating reconstruction. Second, because the weights lie on the simplex and the maps are affine, the composition is itself a well-defined affine perturbation whose gradient with respect to each map is scaled by its weight (Sec.~\ref{sec:fusion}), giving a smooth, non-degenerate learning signal to every expert that receives traffic. The design question is then not whether to condition, but which evidence should condition, and that is what the next two subsections specify.

\subsection{Shared Reconstruction Backbone}
\label{sec:backbone}

Given a Bayer observation, the network predicts $\pred=f_\theta(\raw)$ from a sparse tensor $\raw\in\mathbb{R}^{3\times H\times W}$. Following the alignment argument of RMFA-Net~\cite{li2024rmfa}, we place R, G, and B samples on full-resolution planes rather than packing the four Bayer phases into half-resolution channels, which preserves each sample's sensor-grid coordinate and keeps both green phases on one plane.
This layout matches what ISP reconstruction needs. Brightness and color are properties of the whole scene and are best estimated from a context that has seen the entire frame, whereas texture and edges are local and need fine spatial resolution. A U-Net supplies both: the stride-2 stages build the global context that drives the conditional module, and the shallow stages together with the additive skips carry fine detail back to the output. The encoder (Fig.~\ref{fig:architecture}(a)) extracts multi-scale features through a width-16 stem followed by three stride-2 stages with 32, 64, and 128 channels and two residual blocks per stage. Strided downsampling expands the receptive field while additive skips carry fine-scale structure to the decoder. Two shared bottleneck residual blocks produce $\feat\in\mathbb{R}^{128\times h\times w}$ at one-eighth input resolution---a $32\times32$ grid for a $256\times256$ input.

The decoder mirrors this structure with three stages of convolution, bilinear resizing to the matched encoder grid, skip addition, and residual refinement at widths 128, 64, and 32, the first stage preserving bottleneck resolution. A convolution followed by a $2\times$ PixelShuffle head restores full-resolution RGB. Only the bottleneck is conditional; every mixture shares the same decoder, skips, and head.

\subsection{Local--Global Conditioning}
\label{sec:conditioning}

At each of the $N=hw$ bottleneck locations, a depthwise $3\times3$ convolution supplies local evidence and global average pooling supplies a scene descriptor. Their concatenation forms the conditioning vector:
\begin{align}
\mathbf{u}_i &= \operatorname{DWConv}_{3\times3}(\feat)_i,\\
\mathbf{g} &= \frac{1}{N}\sum_{j=1}^{N}\feat_j,\qquad
\ctx_i=[\mathbf{u}_i;\mathbf{g}]\in\mathbb{R}^{2C}.
\label{eq:context}
\end{align}
The same $\ctx_i$ drives the router and the experts, and that coupling is deliberate: global context can change the transformation itself, not merely the weights assigned to locally predicted candidates. Two locations sharing $\mathbf{g}$ still receive different transformations through their $\mathbf{u}_i$, so spatial adaptation has two independent sources---the mixing weights and the expert outputs. Perturbing either one alone does not remove the other, a property we exploit in Sec.~\ref{sec:routing}.

All $N$ locations are evaluated, giving 1{,}024 tokens for a $256\times256$ input, with no further downsampling inside the module.

\subsection{Affine Experts and Dense Soft Fusion}
\label{sec:fusion}

A linear router predicts a distribution over $E$ experts at each location,
\begin{equation}
 p_{i,k}=\frac{\exp(\mathbf{w}_k^\top\ctx_i+b_k)}
 {\sum_{\ell=1}^{E}\exp(\mathbf{w}_\ell^\top\ctx_i+b_\ell)},
\label{eq:gate}
\end{equation}
and each expert is a one-hidden-layer MLP with separate scale and shift heads,
\begin{align}
\mathbf{h}_{i,k}&=\operatorname{GELU}(A_k\ctx_i+\mathbf{a}_k),\\
\boldsymbol\gamma_{i,k}&=G_k\mathbf{h}_{i,k}+\mathbf{g}_k,\quad
\boldsymbol\beta_{i,k}=B_k\mathbf{h}_{i,k}+\mathbf{b}_k.
\end{align}
The mixture first fuses the parameters and then modulates the shared feature,
\begin{align}
\bar{\boldsymbol\gamma}_i&=\sum_k p_{i,k}\boldsymbol\gamma_{i,k},\quad
\bar{\boldsymbol\beta}_i=\sum_k p_{i,k}\boldsymbol\beta_{i,k},\\
\feat_i'&=\feat_i\odot(\mathbf{1}+\bar{\boldsymbol\gamma}_i)
             +\bar{\boldsymbol\beta}_i.
\label{eq:film}
\end{align}
Because the weights sum to one, fusing parameters is equivalent to mixing the transformations themselves,
\begin{equation}
\feat_i'=\sum_k p_{i,k}\big[\feat_i\odot(\mathbf{1}+\boldsymbol\gamma_{i,k})
+\boldsymbol\beta_{i,k}\big],
\label{eq:equivalence}
\end{equation}
so the model interpolates continuously between processing modes and never commits to a discrete expert. The equivalence holds before the nonlinear decoder, not between independently reconstructed RGB outputs. Although each term is affine in $\feat_i$ for fixed coefficients, both the coefficients and the weights depend on the input, so the module as a whole is not a fixed affine map.

Holding the gate values fixed, the local derivatives with respect to expert outputs are
\begin{equation}
\frac{\partial\feat_i'}{\partial\boldsymbol\beta_{i,k}}=p_{i,k}I,
\qquad
\frac{\partial\feat_i'}{\partial\boldsymbol\gamma_{i,k}}
=p_{i,k}\operatorname{diag}(\feat_i),
\end{equation}
so routing scales the direct learning signal reaching each expert in proportion to the weight it receives. Affine heads are zero-initialized and expert trunks randomly initialized, making Eq.~\eqref{eq:film} the identity at initialization; training proceeds from a stable, unbiased starting point. A $1\times1$ projection then maps the modulated feature from 128 to 256 channels without further spatial reduction.

\subsection{Cost of the Conditional Module}
\label{sec:complexity}

We price the module against the identical architecture with it removed; encoder, bottleneck residual blocks, projection, decoder, and skips are unchanged. For feature width $C$, expert hidden width $d$, and $E$ experts, the added parameters are
\begin{equation}
\Delta P=E(4Cd+d+2C)+E(2C+1)+10C,
\label{eq:complexity}
\end{equation}
covering the affine experts, the linear router, and the depthwise context convolution with biases. With $C=128$, $d=64$, and $E=8$ this is 268{,}040 parameters---264{,}704 in the experts, 2{,}056 in the router, 1{,}280 in local context---a 19.1\% increase over the 1{,}401{,}836-parameter backbone, for a 1.670\,M model (measured total 1{,}669{,}876). The additional convolutional and linear work is $N(4ECd+2CE+9C)$ multiply--accumulates, about 0.272\,G for a $256\times256$ input, excluding pooling, activations, and elementwise fusion. Cost grows linearly in $E$ because every expert is active, and Sec.~\ref{sec:efficiency} reports measured latency for the complete network rather than extrapolating from these counts.

\section{Experiments}
\label{sec:experiments}

Our experiments answer three questions in order: how the model compares with reported results on the target benchmark (Sec.~\ref{sec:results}), whether the mechanism we claim is the one actually responsible (Secs.~\ref{sec:ablation} and~\ref{sec:mixture}), and what that mechanism costs when it runs (Sec.~\ref{sec:efficiency}). The second and third occupy most of our analysis.

\subsection{Datasets and Implementation}
\label{sec:protocol}

\paragraph{Data.}
Our quantitative study uses Fujifilm UltraISP, which pairs mobile-sensor Bayer RAW with Fujifilm RGB references~\cite{ignatov2022pynetv2,ignatov2022microisp}. Its training, development, and evaluation splits contain 23{,}661, 500, and 2{,}258 pairs; the evaluation split is the dataset-provided validation list, referred to here as the test set. The 500-pair development split selects checkpoints by RGB PSNR, and the selected checkpoints are evaluated on the test set with tiling disabled. For cross-sensor evidence we use Zurich RAW to RGB (ZRR), which pairs Huawei P20 RAW with Canon 5D Mark IV photographs at full sensor resolution~\cite{ignatov2020pynet}; it provides 1{,}204 aligned $448\times448$ test patches, and its unaligned full-resolution captures are used only for the qualitative panels of Sec.~\ref{sec:zrr_fullres}. A COCO-derived set~\cite{lin2014coco,xing2021invertible} supplies a further stress test whose RAW inputs come from running InvISP in reverse; we train on it and report reconstruction on 4{,}997 val2017 images. The three models are trained separately.

\paragraph{Preprocessing.}
We subtract the black-level pedestal before normalization, which avoids biasing shadows and color reconstruction~\cite{li2024rmfa}. For the stored RAW10 inputs of Fujifilm UltraISP,
\begin{equation}
\bar r=\frac{\clip((r_{16}\!\gg\!2)-64,\,0,\,959)}{959},
\label{eq:raw}
\end{equation}
where the right shift recovers 10-bit samples under this dataset's storage convention, 64 is the black level, and $959=1023-64$ is the signal range. RGGB samples populate their full-grid R, G, and B planes with zeros at unobserved locations, RGB targets are decoded and divided by 255, and the output head uses ReLU with predictions clamped to $[0,1]$ at evaluation.

\paragraph{Objective and optimization.}
We combine Charbonnier reconstruction, SSIM~\cite{wang2004ssim}, and VGG-based LPIPS~\cite{zhang2018lpips} with a brightness and color-balance term and output regularization,
\begin{equation}
\mathcal{L}=\mathcal{L}_{\mathrm{charb}}
+0.1\mathcal{L}_{\mathrm{ssim}}
+0.1\mathcal{L}_{\mathrm{lpips}}
+\mathcal{L}_{\mathrm{bright}}
+0.1\mathcal{L}_{\mathrm{reg}}.
\label{eq:loss}
\end{equation}
Training uses AdamW in FP32 on eight GPUs with global batch 96, initial backbone learning rate $3\times10^{-4}$, a $2\times$ router rate, weight decay $5\times10^{-4}$, and gradient clipping at 1, with five warmup epochs and cosine decay to $10^{-7}$. Patches are stored at $256\times256$ with no online cropping or augmentation.

\subsection{Comparison with Reported Results}
\label{sec:results}

\begin{table}[t]
\centering\small
\setlength{\tabcolsep}{6pt}
\begin{tabular}{lrr}
\toprule
Method & PSNR $\uparrow$ & SSIM $\uparrow$ \\
\midrule
MiAlgo & 24.19 & 0.8587 \\
AVC2 & 24.18 & 0.8582 \\
UAIC\_ISP & 24.00 & 0.8625 \\
MFI & 23.76 & 0.8564 \\
NAF & 23.69 & 0.8540 \\
amonal & 22.89 & 0.8295 \\
\midrule
RMFA-Net$^{\dagger}$~\cite{li2024rmfa} & 24.377 & 0.87354 \\
\method & \textbf{24.998} & \textbf{0.88366} \\
\bottomrule
\end{tabular}
\caption{Fujifilm UltraISP results (PSNR in dB). The six challenge entries are taken from the MAI 2026 challenge~\cite{mai2026mobileisp}, scored on its hidden test set; $\dagger$ denotes our reimplementation of RMFA-Net.}
\label{tab:sota}
\end{table}

Table~\ref{tab:sota} compares \method{} with the published MAI 2026 entries and with RMFA-Net~\cite{li2024rmfa}, the architecture our reconstruction path descends from. \method{} reaches 24.998\,dB and 0.88366 SSIM, improving on our reimplementation of RMFA-Net by 0.62\,dB.

\paragraph{Additional baseline.}
RMFA-Net~\cite{li2024rmfa} reaches 24.377\,dB PSNR and 0.87354 luminance SSIM on the test set, compared with 24.998\,dB and 0.88366 for \method.

\paragraph{Cross-sensor transfer.}
On ZRR, the ZRR-trained model reaches 21.249\,dB PSNR and 0.7505 RGB SSIM on the aligned test patches. Because ZRR's RAW and its Canon reference come from different cameras, this number measures cross-sensor mapping; the full-resolution captures are shown as whole-frame renderings in Sec.~\ref{sec:zrr_fullres}, without metrics.

\subsection{Zurich RAW to RGB}
\label{sec:zrr}
LG-FiLM and the published PyNET results are compared in Table~\ref{tab:zrr_reference}.

\begin{table}[t]
\centering\small
\setlength{\tabcolsep}{3pt}
\begin{tabular}{lrrr}
\toprule
Method & PSNR (dB) & SSIM (Y) & MS-SSIM (Y) \\
\midrule
PyNET (published)~\cite{ignatov2020pynet} & 21.19 & n/r & 0.8620 \\
\method & 21.249 & 0.7601 & 0.8669 \\
\bottomrule
\end{tabular}
\caption{Reconstruction results on the ZRR test set.}
\label{tab:zrr_reference}
\end{table}

\begin{figure}[t]
\centering
\setlength{\tabcolsep}{2pt}
\begin{tabular}{@{}cccc@{}}
\includegraphics[width=0.235\textwidth]{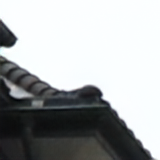} &
\includegraphics[width=0.235\textwidth]{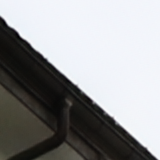} &
\includegraphics[width=0.235\textwidth]{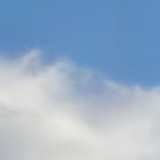} &
\includegraphics[width=0.235\textwidth]{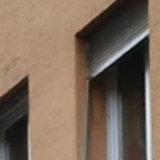} \\[2pt]
\includegraphics[width=0.235\textwidth]{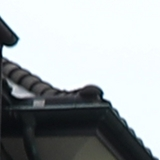} &
\includegraphics[width=0.235\textwidth]{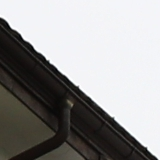} &
\includegraphics[width=0.235\textwidth]{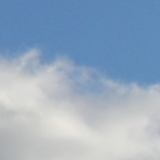} &
\includegraphics[width=0.235\textwidth]{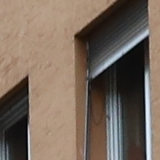} \\
\end{tabular}
\caption{ZRR reconstructions at identical locations in both rows. Top: \method{} outputs; bottom: Canon reference. Patches 1148, 893, 1126 and 879 (left to right) score 32.729/0.9897, 30.738/0.9665, 30.162/0.9573 and 30.840/0.9658 in per-patch RGB PSNR and luminance SSIM.}
\label{fig:zrr_qualitative}
\end{figure}

\paragraph{Results.}
\method{} reaches 21.249\,dB PSNR, 0.7601 SSIM, and 0.8669 MS-SSIM; Fig.~\ref{fig:zrr_qualitative} shows consistent colour and detail reproduction.

\subsection{Full-Resolution Qualitative Comparison}
\label{sec:zrr_fullres}

Figures~\ref{fig:zrr_full_1503}--\ref{fig:zrr_full_1050} show full-resolution ZRR reconstructions and modulation comparisons. Exposure, white balance, and gradients in sky and foliage remain plausible at native resolution. Full-resolution pairs are unaligned and shown qualitatively.

\newcommand{\zrrfull}[3]{%
\begin{figure}[t]
\centering
\begin{tabular}{@{}cc@{}}
\includegraphics[width=0.48\textwidth]{#1_gt.jpg} &
\includegraphics[width=0.48\textwidth]{#1_ours.jpg} \\
{\small (a) Canon reference} & {\small (b) \method} \\[2pt]
\includegraphics[width=0.48\textwidth]{#1_bypass.jpg} &
\includegraphics[width=0.48\textwidth]{#1_prior.jpg} \\
{\small (c) Modulation bypassed} & {\small (d) Fixed routing prior} \\
\end{tabular}
\caption{Full-resolution ($3968\times2976$) ZRR reconstruction. (a) Canon reference; (b) \method; (c) modulation bypassed; (d) fixed prior. #2}
\label{#3}
\end{figure}
}

\zrrfull{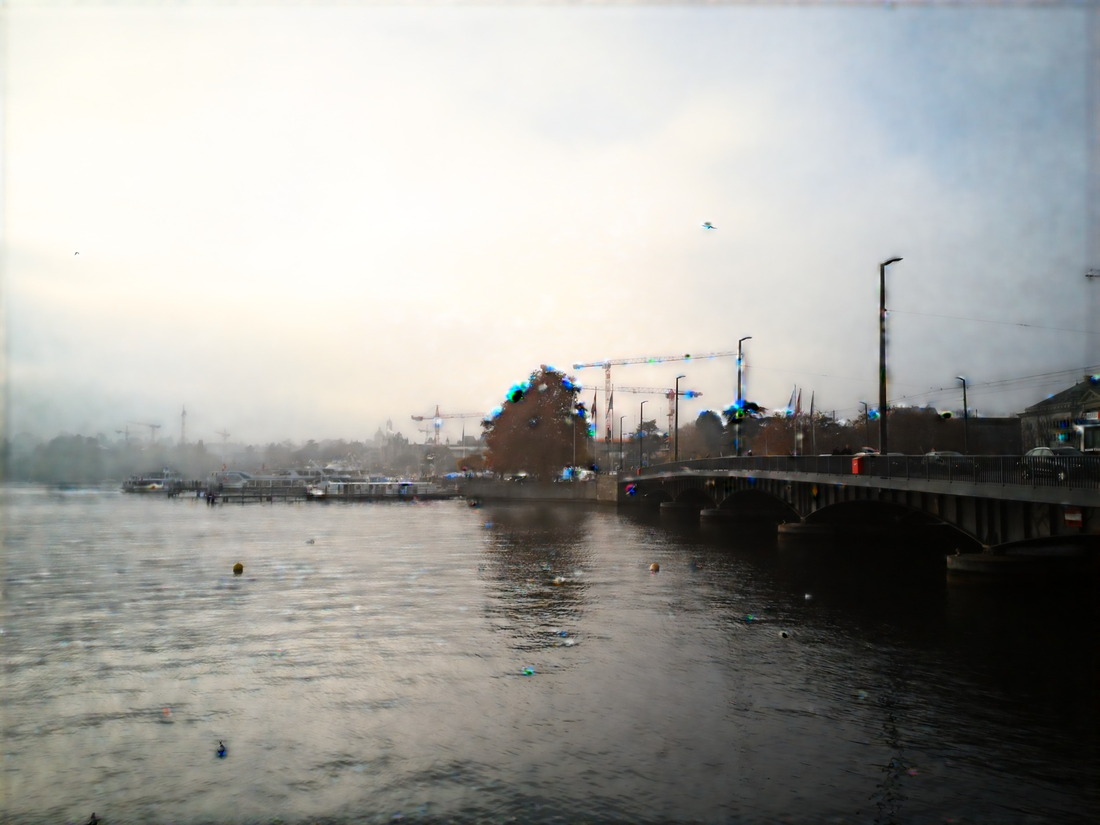}{Broad scene coverage and a wide tonal range.}{fig:zrr_full_1503}

\zrrfull{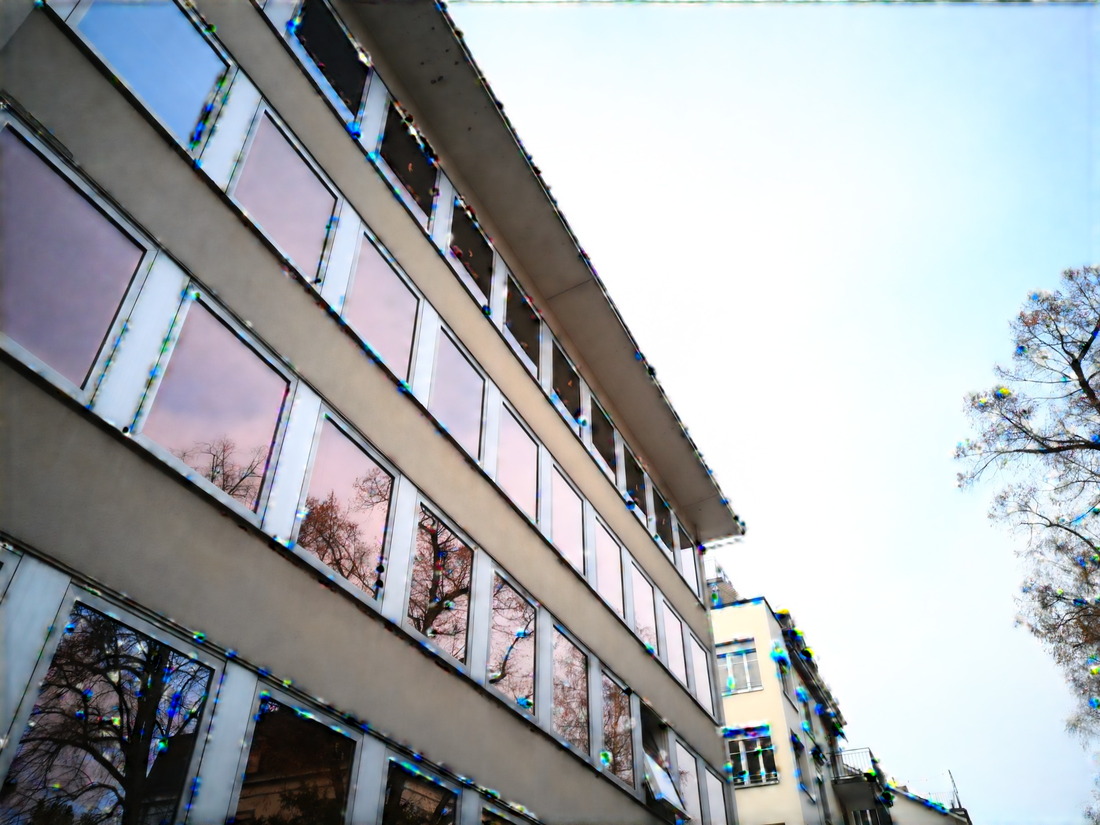}{A bright sky above a deeply shaded foreground.}{fig:zrr_full_1909}

\zrrfull{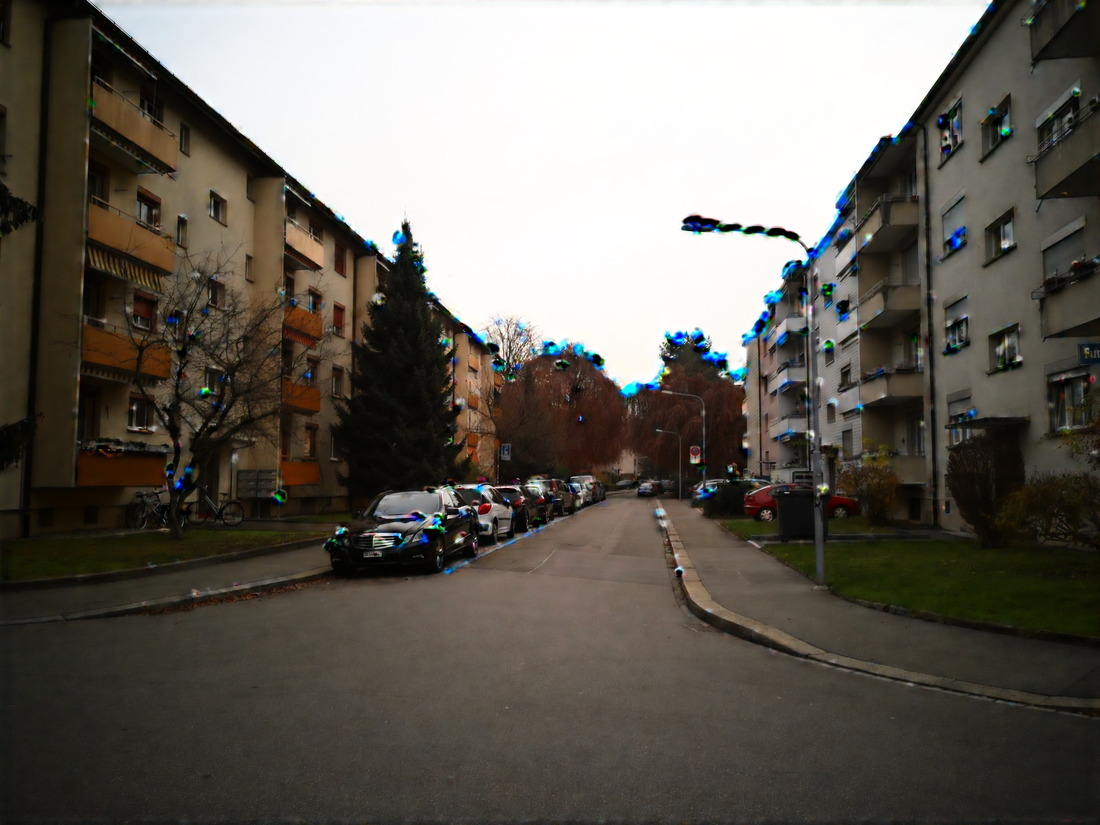}{A street scene with vegetation and fine structural detail.}{fig:zrr_full_4664}

\zrrfull{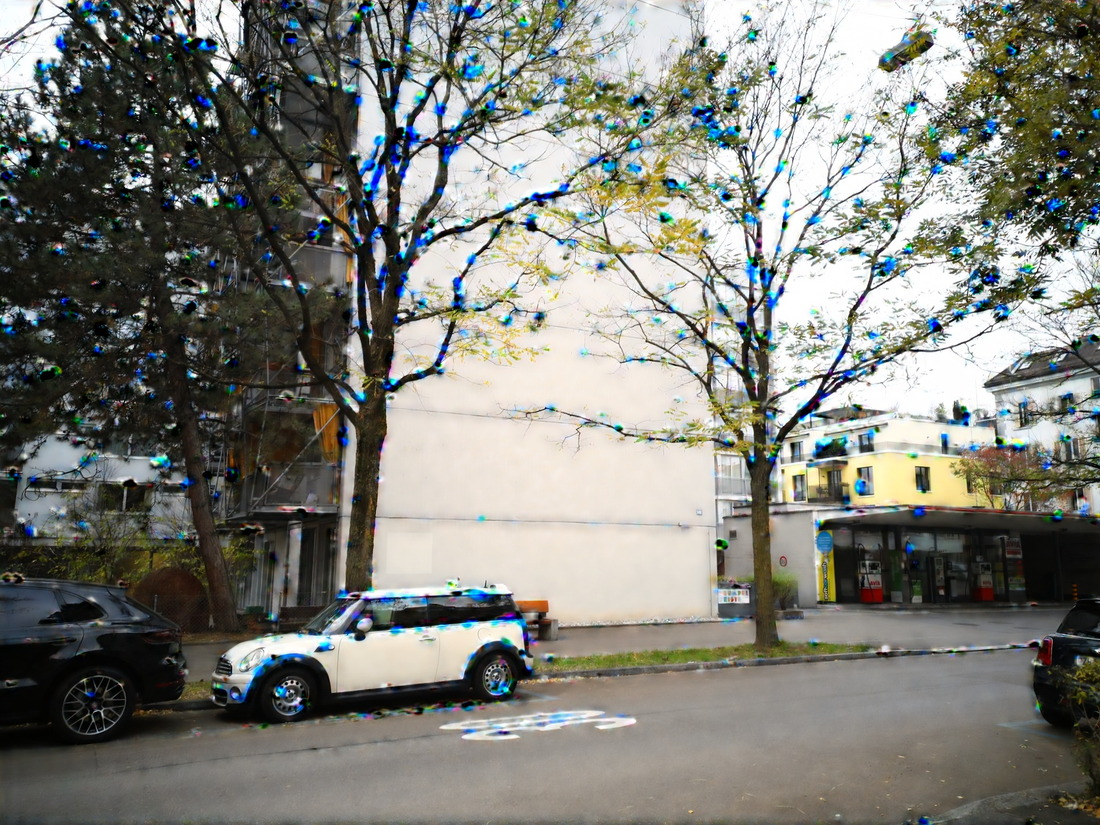}{A further test scene.}{fig:zrr_full_1050}

\paragraph{Derived-RAW third domain.}
Retrained from scratch on the InvISP-derived COCO data, the same architecture reaches 35.802\,dB PSNR and 0.9815 SSIM on the val2017 test images (Sec.~\ref{sec:coco}), evidence that it reconstructs a structurally different RAW domain.

\subsection{InvISP-Derived COCO}
\label{sec:coco}

We evaluate \method{} on COCO RAW--RGB pairs synthesized with InvISP~\cite{xing2021invertible} from photographs~\cite{lin2014coco}.

\begin{table}[t]
\centering\small
\setlength{\tabcolsep}{4pt}
\begin{tabular}{lr}
\toprule
Quantity & Value \\
\midrule
PSNR (RGB) & 35.802 \\
SSIM (Y) & 0.9815 \\
MS-SSIM (Y) & 0.9976 \\
SSIM (RGB) & 0.9624 \\
\bottomrule
\end{tabular}
\caption{Reconstruction results on the InvISP-derived COCO test set.}
\label{tab:coco}
\end{table}

\begin{figure}[t]
\centering
\setlength{\tabcolsep}{2pt}
\begin{tabular}{@{}cccc@{}}
\includegraphics[width=0.23\textwidth]{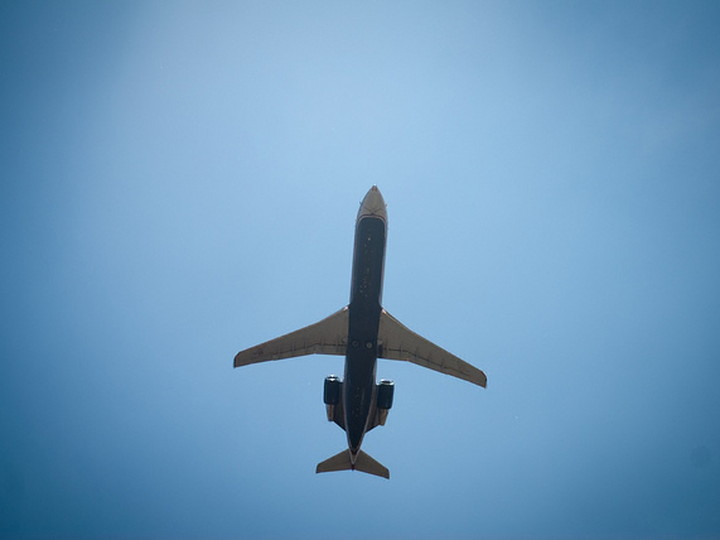} &
\includegraphics[width=0.23\textwidth]{000000491757_pred.jpg} &
\includegraphics[width=0.23\textwidth]{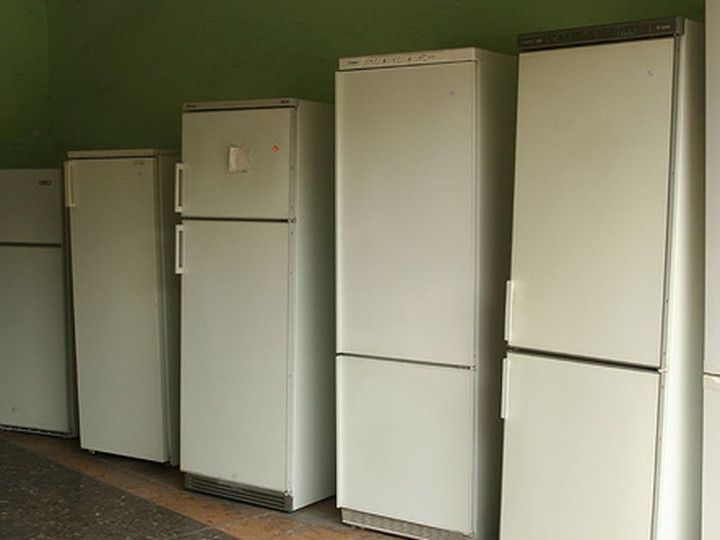} &
\includegraphics[width=0.23\textwidth]{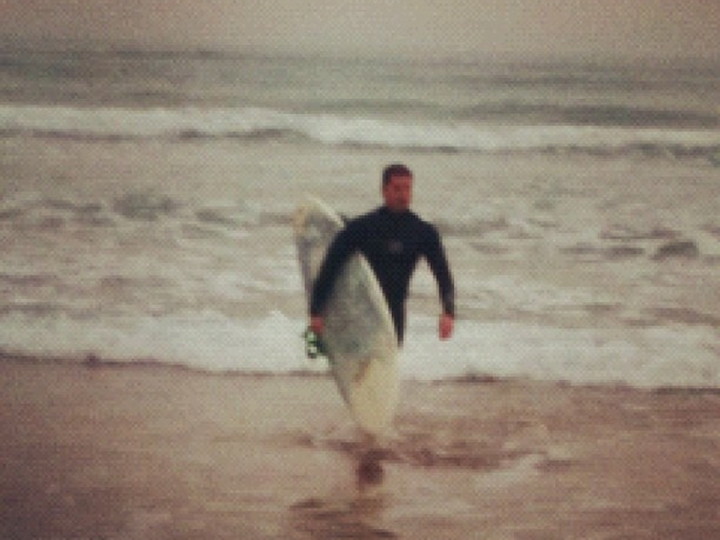} \\[2pt]
\includegraphics[width=0.23\textwidth]{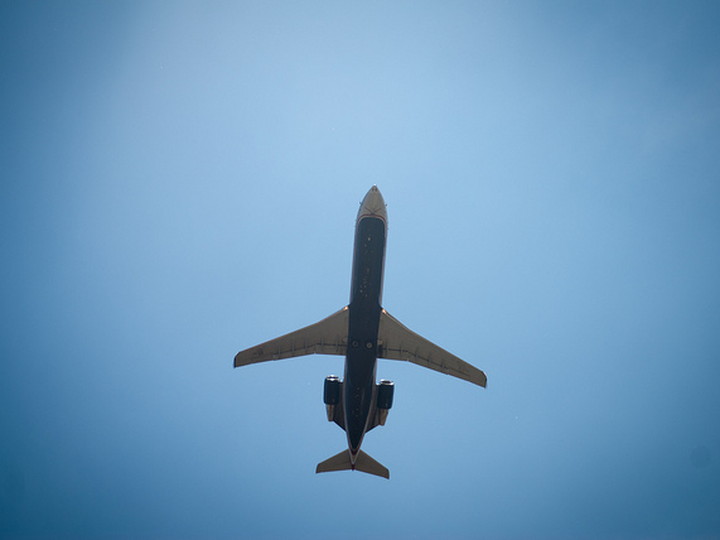} &
\includegraphics[width=0.23\textwidth]{000000491757_gt.jpg} &
\includegraphics[width=0.23\textwidth]{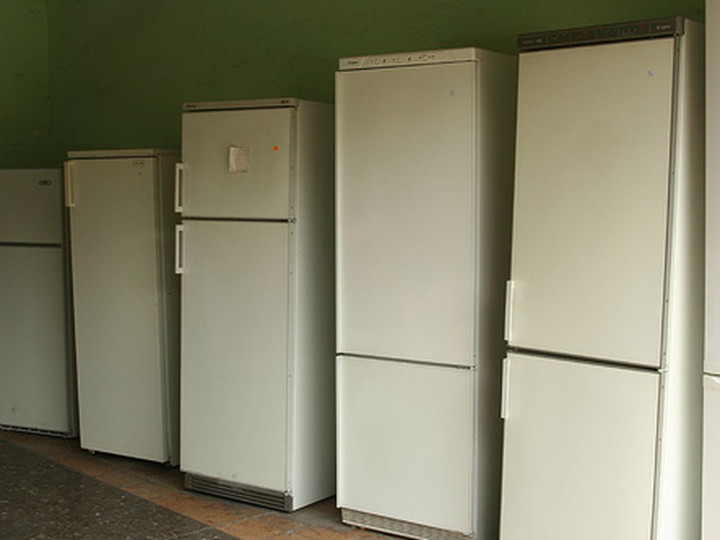} &
\includegraphics[width=0.23\textwidth]{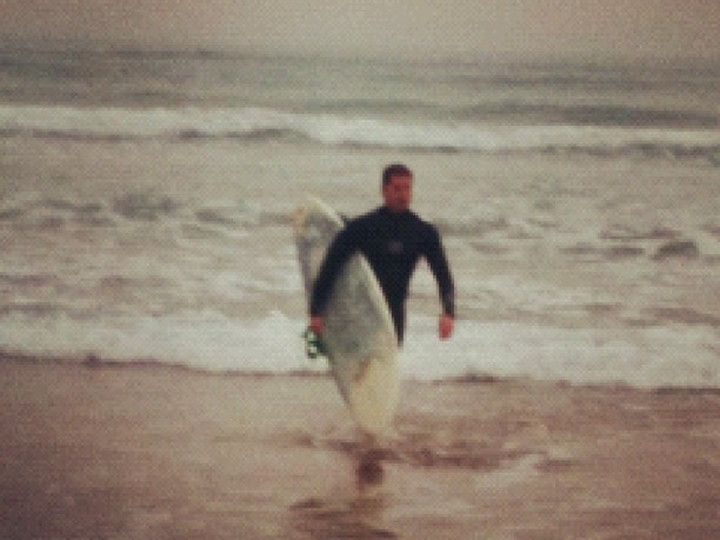} \\
\end{tabular}
\caption{Reconstruction examples on the InvISP-derived COCO test set. Top row: \method{} outputs; bottom row: the COCO references.}
\label{fig:coco}
\end{figure}

\paragraph{Results.}
Table~\ref{tab:coco} summarizes reconstruction accuracy, and Fig.~\ref{fig:coco} shows visual examples. Together, they demonstrate the model's ability to learn another RAW-to-RGB mapping.

\subsection{Ablation Study}
\label{sec:ablation}

\subsection{Architectural Components}
\label{sec:architecture_supp}
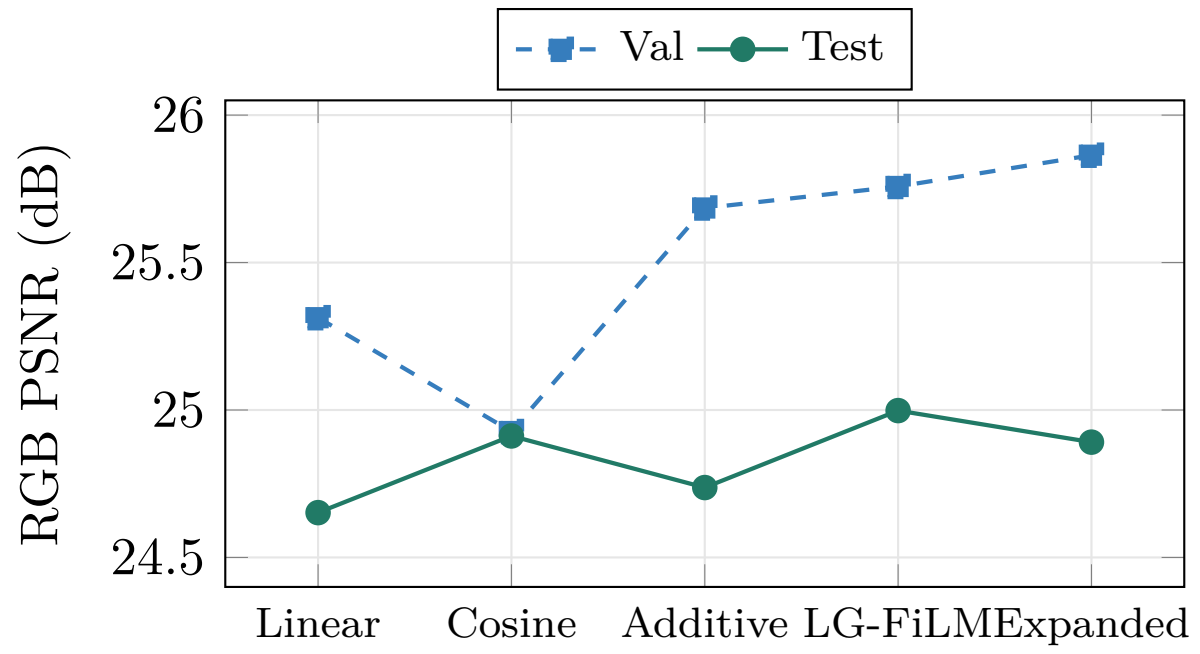
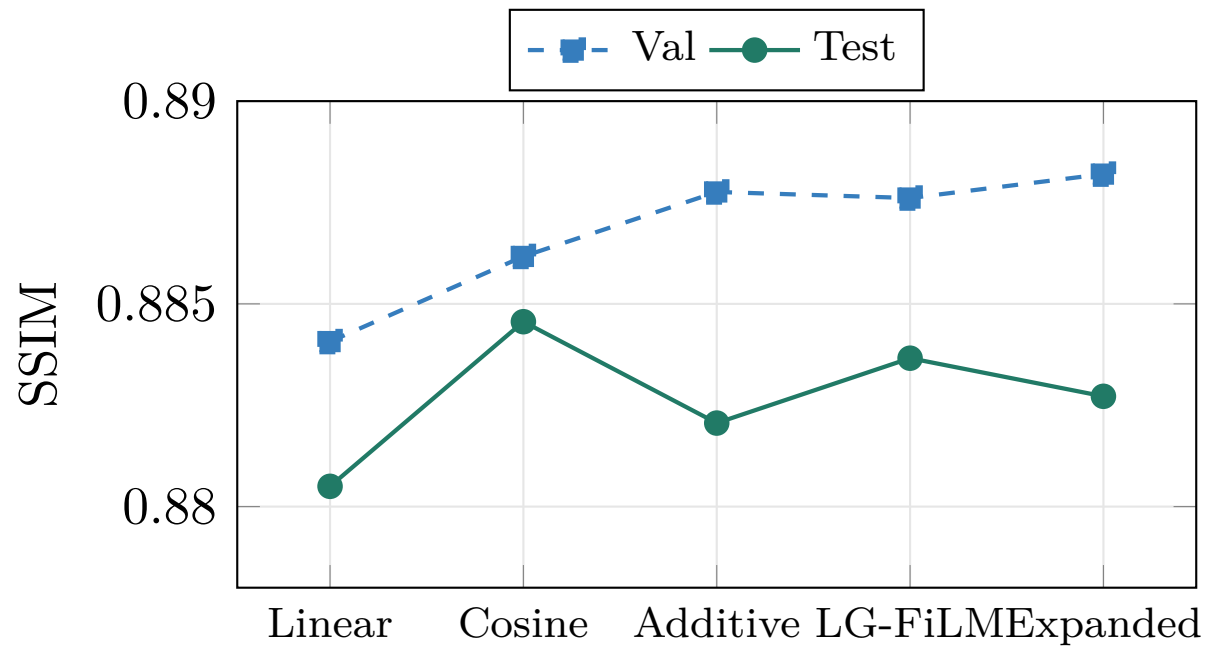
\begin{figure}[t]
\centering
\begin{subfigure}{0.48\textwidth}
\centering
\begin{tikzpicture}
\begin{axis}[width=\linewidth,height=4.8cm,
 ylabel={RGB PSNR (dB)},
 symbolic x coords={Linear,Cosine,Additive,LG-FiLM,Expanded},xtick=data,
 xticklabel style={font=\scriptsize},
 ymin=24.4,ymax=26.05,grid=major,grid style={gray!20},
 tick label style={font=\small},label style={font=\small},
 legend style={font=\scriptsize,at={(0.5,1.02)},anchor=south,legend columns=2},
 enlarge x limits=0.12]
\addplot[color=plotblue,mark=square*,thick,dashed] coordinates {
 (Linear,25.313) (Cosine,24.927) (Additive,25.685) (LG-FiLM,25.758) (Expanded,25.864)};
\addlegendentry{Val}
\addplot[color=plotgreen,mark=*,thick] coordinates {
 (Linear,24.652) (Cosine,24.912) (Additive,24.737) (LG-FiLM,24.998) (Expanded,24.891)};
\addlegendentry{Test}
\end{axis}
\end{tikzpicture}
\caption{PSNR ordering differs between the Val and Test sets.}
\end{subfigure}\hfill
\begin{subfigure}{0.48\textwidth}
\centering
\begin{tikzpicture}
\begin{axis}[width=\linewidth,height=4.8cm,
 ylabel={SSIM},
 symbolic x coords={Linear,Cosine,Additive,LG-FiLM,Expanded},xtick=data,
 xticklabel style={font=\scriptsize},
 ymin=0.878,ymax=0.890,grid=major,grid style={gray!20},
 scaled y ticks=false,yticklabel style={/pgf/number format/fixed,/pgf/number format/precision=3},
 tick label style={font=\small},label style={font=\small},
 legend style={font=\scriptsize,at={(0.5,1.02)},anchor=south,legend columns=2},
 enlarge x limits=0.12]
\addplot[color=plotblue,mark=square*,thick,dashed] coordinates {
 (Linear,0.88407) (Cosine,0.88617) (Additive,0.88776) (LG-FiLM,0.88761) (Expanded,0.88820)};
\addlegendentry{Val}
\addplot[color=plotgreen,mark=*,thick] coordinates {
 (Linear,0.88050) (Cosine,0.88456) (Additive,0.88206) (LG-FiLM,0.88366) (Expanded,0.88272)};
\addlegendentry{Test}
\end{axis}
\end{tikzpicture}
\caption{SSIM ordering differs from PSNR.}
\end{subfigure}
\caption{Local quality on the Val and Test sets. Linear/Cosine denote image-level routers, Additive denotes local--global additive experts, and Expanded denotes the conditioned 16-expert variant. Expanded improves Val but falls below \method{} on Test.}
\label{fig:results}
\end{figure}

The comparisons examine the information available for routing, the form and capacity of expert transformations, and the shared reconstruction path. Tables~\ref{tab:main}--\ref{tab:ablation} and Figure~\ref{fig:results} report the individual design changes, and Sec.~\ref{sec:sweep_budget} retrains the main conditioning variants side by side and reports them against the main model in Table~\ref{tab:sweep_budget}.

\begin{table*}[t]
\centering\small
\setlength{\tabcolsep}{6pt}
\begin{tabular}{lrrrrr}
\toprule
Model / configuration & Params (M) & Val PSNR & Val SSIM & Test PSNR & Test SSIM \\
\midrule
Image-level linear & 2.956 & 25.313 & 0.88407 & 24.652 & 0.88050 \\
Image-level cosine & 2.960 & 24.927 & 0.88617 & 24.912 & 0.88456 \\
Image-level soft & -- & 24.626 & 0.87852 & 24.507 & 0.87717 \\
NAF-style ISP (local) & -- & 25.441 & \textbf{0.89067} & 24.681 & \textbf{0.88618} \\
Additive tokens & 2.491 & 25.685 & 0.88776 & 24.737 & 0.88206 \\
No conditioning & 1.402 & 23.846 & 0.87760 & 23.366 & 0.87430 \\
\textbf{\method} & \textbf{1.670} & 25.758 & 0.88761 & \textbf{24.998} & 0.88366 \\
Conditional, $E=16$ & 2.018 & \textbf{25.864} & 0.88820 & 24.891 & 0.88272 \\
\bottomrule
\end{tabular}
\caption{Architectural comparisons on the Val and Test sets. ``No conditioning'' is the same encoder--decoder with the bottleneck module removed; bold marks the best value per column.}
\label{tab:main}
\end{table*}

\paragraph{Routing granularity.}
Image-level linear, cosine, and soft gates assign one expert mixture to the entire image, whereas local--global token conditioning varies transformations with spatial content while sharing scene context. In Table~\ref{tab:main}, \method{} improves Test PSNR over the linear and cosine variants by 0.346 and 0.086\,dB; the corresponding Val gains are 0.445 and 0.831\,dB. Sec.~\ref{sec:expert_analysis} examines the spatial contribution of routing weights.

\begin{table}[t]
\centering\small
\setlength{\tabcolsep}{3pt}
\begin{tabular}{lrrr}
\toprule
Change from \method & Params (M) & PSNR & $\Delta$ \\
\midrule
None & 1.670 & 25.758 & -- \\
Additive experts & 2.491 & 25.685 & $-0.073$ \\
Remove skips & 1.670 & 25.237 & $-0.521$ \\
Rep. training blocks & 3.027 & 25.754 & $-0.004$ \\
Decoder conditioning & 1.745 & 25.748 & $-0.010$ \\
Conditioning + $E=16$ & 2.018 & 25.864 & $+0.106$ \\
VSS-inspired blocks & -- & 25.506 & $-0.252$ \\
\bottomrule
\end{tabular}
\caption{RGB PSNR on the Val set for architectural changes. $\Delta$ subtracts the displayed score; reparameterized parameters are counted before fusion, and the expanded variant jointly adds RAW token conditioning and raises the expert count to $E=16$.}
\label{tab:ablation}
\end{table}

\paragraph{Skip connections.}
Removing additive encoder--decoder skips routes all information through the modulated bottleneck; Val PSNR falls by 0.521\,dB.

\paragraph{Expert formulation.}
Additive experts cannot adjust response magnitude. The affine form improves PSNR by 0.073\,dB on Val and 0.261\,dB on Test while reducing parameters from 2.491 to 1.670\,M; expert width changes together with output form.

\paragraph{Reconstruction blocks.}
Reparameterized convolutions, local NAF-style blocks, and VSS-inspired directional processing test richer feature extraction. Reparameterization scores 25.754\,dB against 25.758\,dB for \method, while the NAF-style and VSS-inspired variants reach 25.441 and 25.506\,dB. NAF-style processing has the highest SSIM in Table~\ref{tab:main}, and \method{} has the highest PSNR on both sets.

\paragraph{Decoder conditioning.}
Adding scene features and RAW statistics to several decoder stages gives 25.748\,dB against 25.758\,dB for \method{} with 1.745\,M rather than 1.670\,M parameters, so repeated conditioning brings no observed gain over bottleneck-only modulation.

\paragraph{Additional training records.}
Table~\ref{tab:history} collects the additional Val records available for these variants.

\begin{table*}[t]
\centering\small
\setlength{\tabcolsep}{6pt}
\begin{tabular}{lrr}
\toprule
Configuration & RGB PSNR & SSIM \\
\midrule
RAW token conditioning + 16 experts & \textbf{25.864} & 0.88820 \\
LG-FiLM (local--global affine mixture) & 25.758 & 0.88761 \\
Reparameterized training blocks & 25.754 & -- \\
Additional decoder conditioning & 25.748 & -- \\
Local--global additive experts & 25.685 & 0.88776 \\
VSS-inspired spatial blocks & 25.506 & 0.88394 \\
NAF-style (linear-baseline derivative) & 25.441 & \textbf{0.89067} \\
Image-level linear router & 25.313 & 0.88407 \\
NAF-style (cosine-baseline derivative) & 25.245 & 0.88581 \\
Initialized expert pool + cosine router & 25.240 & 0.88420 \\
Remove additive skips & 25.237 & -- \\
Image-level cosine router & 24.927 & 0.88617 \\
GAP-only multilayer router & 24.923 & 0.88100 \\
Image-level soft gate & 24.626 & 0.87852 \\
GAP/GMP multilayer router & 24.588 & 0.88089 \\
No conditioning (bottleneck module removed) & 23.846 & 0.87760 \\
\bottomrule
\end{tabular}
\caption{Additional model comparisons on Val, sorted by PSNR. Dashes indicate unreported values.}
\label{tab:history}
\end{table*}

\paragraph{Quantitative comparison.}
Table~\ref{tab:deltas} uses displayed mean PSNR values and unrounded parameter counts. Parameter reduction is measured relative to the comparison model:
\begin{equation}
R_P=100\left(1-\frac{P_{\mathrm{LG\text{-}FiLM}}}{P_{\mathrm{reference}}}\right).
\end{equation}
The expanded variant adds 20.8\% parameters, whereas choosing \method{} saves 17.2\%.

\begin{table}[t]
\centering\small
\setlength{\tabcolsep}{3pt}
\begin{tabular}{lrr}
\toprule
Reference & LG-FiLM $-$ ref. (dB) & Fewer params \\
\midrule
Image-level linear & $+0.445$ & 43.5\% \\
Image-level cosine & $+0.831$ & 43.6\% \\
Additive tokens & $+0.073$ & 33.0\% \\
Expanded variant & $-0.106$ & 17.2\% \\
\bottomrule
\end{tabular}
\caption{Derived trade-offs on the Val set. Positive PSNR differences favor \method; positive parameter reductions indicate a smaller model.}
\label{tab:deltas}
\end{table}

\paragraph{Number of experts.}
An expanded variant that adds decoder conditioning and RAW-statistic expert inputs raises parameters from 1.670 to 2.018\,M but lowers Test PSNR/SSIM from 24.998/0.88366 to 24.891/0.88272, while improving Val from 25.758/0.88761 to 25.864/0.88820. Sec.~\ref{sec:scaling_probe} reports a controlled sweep over $E\in\{1,2,4,16\}$ that varies only the expert count; none reaches the eight-expert model on Test.

\paragraph{Routing strategy.}
The main table evaluates the soft mixture, token-hard argmax, and a target-assisted single-expert oracle on the Test set; Table~\ref{tab:oracle_val} repeats the analysis on the Val set.

\begin{table}[t]
\centering\small
\setlength{\tabcolsep}{2pt}
\begin{tabular}{lrrrr}
\toprule
Model & Soft & Token-hard & Oracle & Oracle$-$soft \\
\midrule
\method & 25.758 & 25.014 & 26.184 & $+0.427$ \\
No skips & 25.237 & 23.076 & 24.313 & $-0.924$ \\
Reparameterized & 25.754 & 24.599 & 26.331 & $+0.578$ \\
Decoder cond. & 25.748 & 25.783 & 27.185 & $+1.436$ \\
Expanded ($E=16$) & 25.864 & 25.673 & 27.441 & $+1.577$ \\
\bottomrule
\end{tabular}
\caption{Routing analysis on the Val set; RGB PSNR in dB.}
\label{tab:oracle_val}
\end{table}

For \method, hard selection costs 0.744\,dB. For the no-skip variant the single-expert oracle is 24.313\,dB, below its soft mixture's 25.237\,dB, so an oracle over single-expert outputs is not a universal upper bound.

\subsection{Expert-count scan}
\label{sec:scaling_probe}

We compare $E\in\{1,2,4,16\}$ with the eight-expert model, which is the main model.

Enlarging the expert bank from 1.437\,M to 1.937\,M parameters ($+34.8\%$) produces no accuracy gain: the four Test PSNR values span 24.780 to 24.856\,dB, a range of 0.076\,dB, and all four fall below the eight-expert model at 24.998\,dB. The SSIM spread is 0.0019 and does not follow the PSNR ordering, so adding experts does not improve reconstruction in this range. This scan varies $E$ alone; Sec.~\ref{sec:sweep_budget} applies the same retrained comparison to the conditioning designs themselves, including the removal of each conditioning input.

\begin{table}[t]
\centering
\footnotesize
\setlength{\tabcolsep}{3pt}
\begin{tabular}{lccccc}
\toprule
$E$ & Params (M) & Val PSNR & Test PSNR & Test SSIM & MS-SSIM \\
\midrule
1 & 1.4365 & 25.452 & 24.856 & 0.8829 & 0.9589 \\
2 & 1.4698 & 25.652 & 24.805 & 0.8833 & 0.9590 \\
4 & 1.5365 & 25.628 & 24.780 & 0.8830 & 0.9590 \\
8 & 1.6699 & 25.758 & \textbf{24.998} & \textbf{0.8837} & \textbf{0.9592} \\
16 & 1.9366 & \textbf{25.760} & 24.816 & 0.8813 & 0.9586 \\
\bottomrule
\end{tabular}
\caption{Expert-count comparison. $E{=}8$ is the main model.}
\label{tab:scaling_probe}
\end{table}

\subsection{Ablations of the main model}
\label{sec:sweep_budget}

Table~\ref{tab:sweep_budget} places the main model beside variants that remove or replace one part of its conditioning design. All rows share one recipe---the same training pairs, batch 12 per process on eight GPUs, initial rate $3\times10^{-4}$, and a configured 4{,}000 epochs---so the differences reflect the architecture. Checkpoints are selected on the development split by RGB PSNR, and the test set is used only for reporting.

\begin{table}[t]
\centering\scriptsize
\setlength{\tabcolsep}{2.5pt}
\begin{tabular}{lrrrr}
\toprule
Variant & Params (M) & Test PSNR & $\Delta$ & Test SSIM \\
\midrule
\textbf{LG-FiLM (main model)} & \textbf{1.670} & \textbf{24.998} & -- & \textbf{0.88366} \\
Eight experts, retrained & 1.670 & 24.817 & $-0.181$ & 0.88235 \\
Single FiLM, local+global & 1.436 & 24.912 & $-0.086$ & 0.88304 \\
Wide single FiLM, local+global & 1.670 & 24.840 & $-0.158$ & 0.88257 \\
Global only, single FiLM & 1.427 & 24.857 & $-0.141$ & 0.88336 \\
Local only & 1.428 & 23.387 & $-1.611$ & 0.87599 \\
No conditioning (backbone) & 1.402 & 23.366 & $-1.632$ & 0.87433 \\
\bottomrule
\end{tabular}
\caption{Ablations of \method{} on the test set. $\Delta$ is the PSNR difference from the main model; SSIM is measured on the Y channel.}
\label{tab:sweep_budget}
\end{table}

Removing the conditioning module costs 1.63\,dB. The two conditioning inputs are not equally important. Keeping only the local branch gives 23.387\,dB, essentially the same as the unmodulated backbone at 23.366\,dB, whereas keeping only the global branch gives 24.857\,dB. A $3\times3$ neighbourhood sees little beyond local texture, while the pooled feature describes the whole scene, and it is the global descriptor that changes the processing mode.

Enlarging the conditional branch does not help. A single FiLM pair with hidden width 520 has the same 1.670\,M parameters as the eight-expert mixture and reaches 24.840\,dB, a difference of 0.02\,dB, and a narrower single pair with 1.436\,M parameters reaches 24.912\,dB. Neither the number of transformations nor their width changes the result on this benchmark.

The second row gives the run-to-run spread for reference. It is the same architecture as the main model, retrained with the same recipe, and it scores 0.181\,dB lower. The four variants that keep both inputs span 24.817 to 24.912\,dB, a range of 0.095\,dB, which is smaller than that spread.

We compare expert formulations, skip connections, and conditioning designs (Table~\ref{tab:architecture_summary}).

\paragraph{The reconstruction path.}
Removing the skip connections routes all information through the bottleneck and costs 0.344\,dB (Table~\ref{tab:architecture_summary}), confirming that the module modulates detail rather than replacing it. Substituting additive experts for multiplicative scale--shift experts costs 0.261\,dB while using 49\% more parameters, so the multiplicative form is preferable here. The remaining rows are controls on where capacity is spent: reparameterized training blocks cost 0.168\,dB, additional decoder conditioning 0.346\,dB, and a VSS-inspired feature extractor 0.334\,dB, while a NAF-style ISP reaches 24.681\,dB.

\begin{table}[t]
\centering\small
\setlength{\tabcolsep}{3pt}
\begin{tabular}{lrrr}
\toprule
Configuration & Params (M) & PSNR & $\Delta$ \\
\midrule
\method & 1.670 & 24.998 & -- \\
No conditioning & 1.402 & 23.366 & $-1.632$ \\
Additive experts & 2.491 & 24.737 & $-0.261$ \\
Remove skips & 1.670 & 24.654 & $-0.344$ \\
Rep.\ training blocks & 3.027 & 24.830 & $-0.168$ \\
Decoder conditioning & 1.745 & 24.652 & $-0.346$ \\
VSS-inspired blocks & -- & 24.664 & $-0.334$ \\
\bottomrule
\end{tabular}
\caption{Architectural ablations (RGB PSNR, dB). No conditioning removes the modulation module. Reparameterized counts are before fusion.}
\label{tab:architecture_summary}
\end{table}

\paragraph{Conditioning granularity.}
Local--global conditioning improves test PSNR over image-level linear and cosine routing by 0.346 and 0.086\,dB. These comparisons change the conditioning design, whereas spatial shuffling in Sec.~\ref{sec:routing} changes only the routing weights.

\paragraph{Expert count.}
We evaluate $E\in\{1,2,4,8,16\}$ (Table~\ref{tab:expert_scan}), where $E{=}8$ is the main model. All four sweep runs finish below the eight-expert model's 24.998\,dB: their test PSNR spans only 24.780 to 24.856\,dB, a spread of 0.076\,dB, and enlarging the bank from $E{=}1$ to $E{=}16$ adds 34.8\% parameters (1.437\,M to 1.937\,M) without closing the gap. Expert count therefore buys no accuracy anywhere in this range, and we keep eight. A decoder-conditioned variant that feeds local RAW statistics to the router and experts and raises the bank to sixteen experts (2.018\,M parameters, +20.8\%) improves the development score from 25.758 to 25.864 but lowers test PSNR to 24.891\,dB and SSIM to 0.88272.

\begin{table}[t]
\centering\small
\setlength{\tabcolsep}{3pt}
\begin{tabular}{lrrrr}
\toprule
$E$ & Params & Val & Test PSNR & Test SSIM \\
 & (M) & PSNR & & \\
\midrule
1 & 1.437 & 25.452 & 24.856 & 0.88290 \\
2 & 1.470 & 25.652 & 24.805 & 0.88326 \\
4 & 1.536 & 25.628 & 24.780 & 0.88303 \\
8 & 1.670 & 25.758 & \textbf{24.998} & \textbf{0.88366} \\
16 & 1.937 & \textbf{25.760} & 24.816 & 0.88134 \\
\bottomrule
\end{tabular}
\caption{Expert-count comparison. Val PSNR is measured on the development split; test scores use the test set.}
\label{tab:expert_scan}
\end{table}

\subsection{What the Mixture Learns}
\label{sec:mixture}

\begin{figure}[t]\centering
\setlength{\tabcolsep}{1.5pt}
\begin{tabular}{@{}cccccc@{}}
\includegraphics[width=0.155\textwidth]{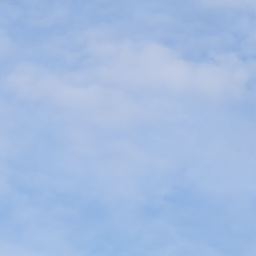} &
\includegraphics[width=0.155\textwidth]{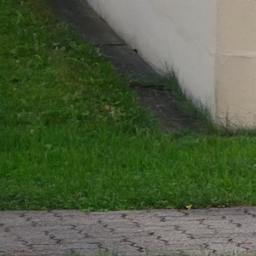} &
\includegraphics[width=0.155\textwidth]{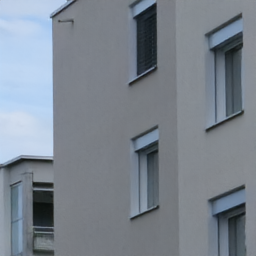} &
\includegraphics[width=0.155\textwidth]{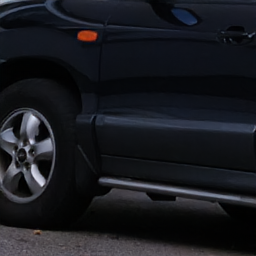} &
\includegraphics[width=0.155\textwidth]{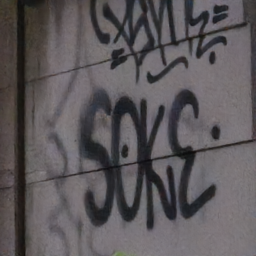} &
\includegraphics[width=0.155\textwidth]{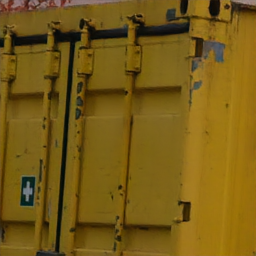} \\[2pt]
\includegraphics[width=0.155\textwidth]{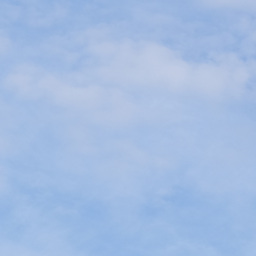} &
\includegraphics[width=0.155\textwidth]{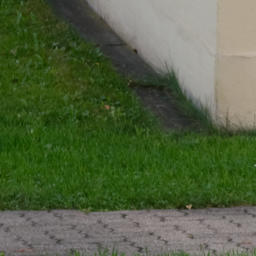} &
\includegraphics[width=0.155\textwidth]{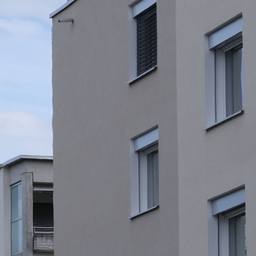} &
\includegraphics[width=0.155\textwidth]{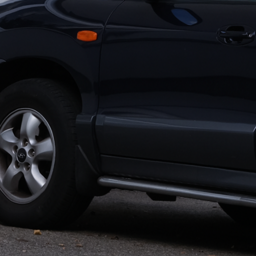} &
\includegraphics[width=0.155\textwidth]{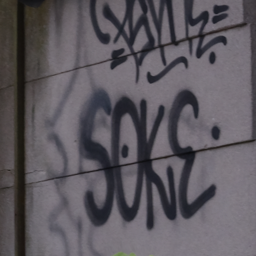} &
\includegraphics[width=0.155\textwidth]{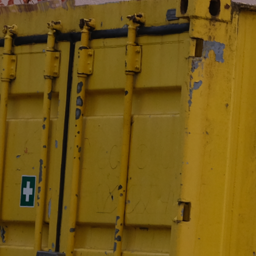} \\[2pt]
\end{tabular}
\caption{Qualitative comparison across varied scenes. Top: \method{} outputs; bottom: the corresponding references.}
\label{fig:qualitative}
\end{figure}

\begin{figure}[t]
\centering
\includegraphics[width=\textwidth]{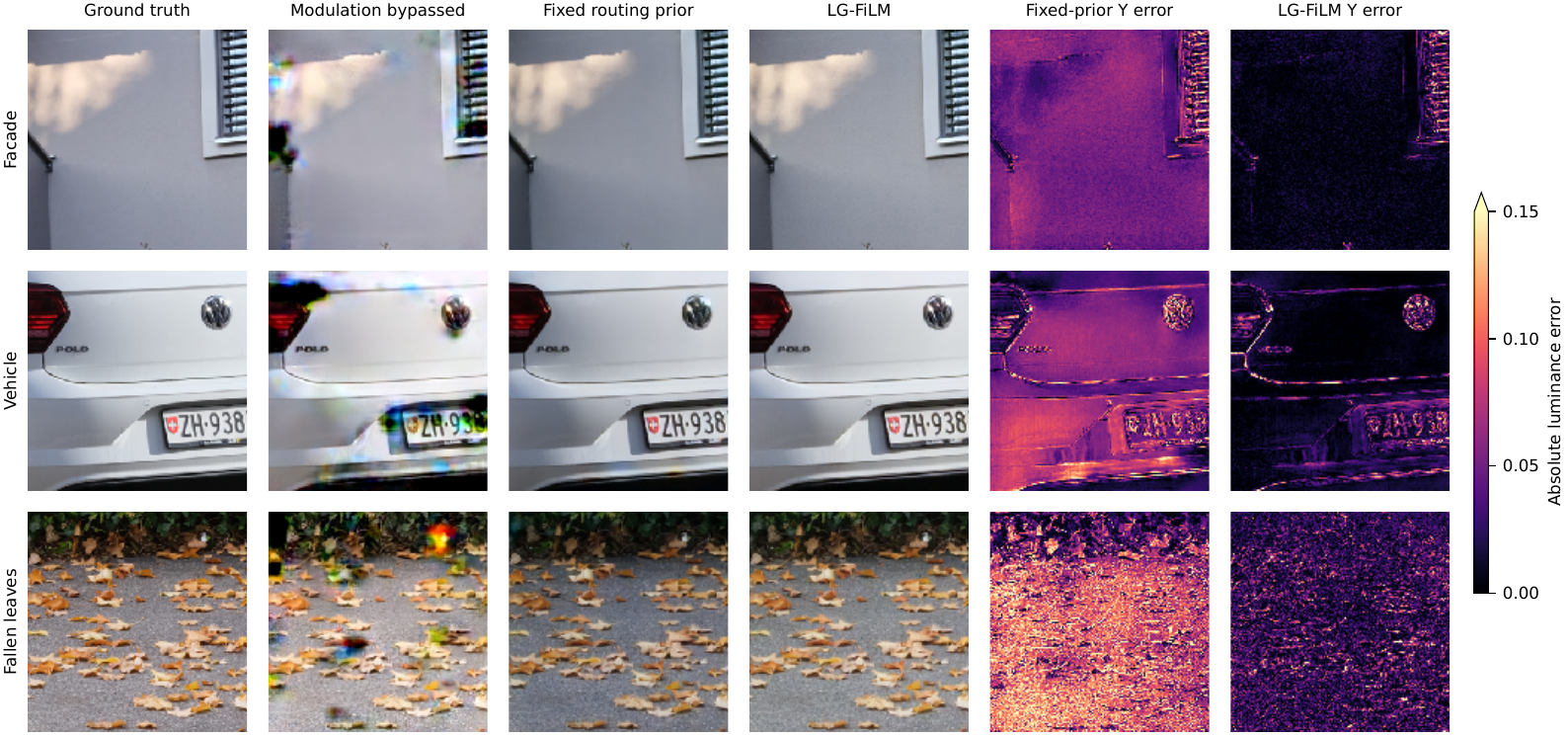}
\caption{Effect of expert modulation and adaptive routing. Three test-set scenes compare GT, bypassed modulation, a fixed routing prior, and the learned mixture. The prior fixes mean weights across inputs and locations but retains input-conditioned expert transformations. RGB panels share a display range; absolute luminance errors share $[0,0.15]$, with larger values saturating.}
\label{fig:mechanism_comparison}
\end{figure}

\paragraph{Controlled interventions and expert contributions.}
\label{sec:expert_analysis}
\paragraph{Controlled interventions.}
All interventions use the same frozen eight-expert checkpoint. Bypassing modulation sets $\gamma=\beta=0$; uniform routing replaces every weight vector with $1/8$; permuted routing cyclically mismatches weights and experts over all seven nonidentity shifts; the fixed prior uses the mean learned weights over the test inputs at every token, without ground truth, and retains input-conditioned expert transformations. Image-mean routing broadcasts each image's mean weights, and spatial shuffling permutes weight vectors within an image under a fixed seed.

\begin{table}[t]
\centering\small
\setlength{\tabcolsep}{3pt}
\begin{tabular}{lrr}
\toprule
Inference intervention & RGB PSNR & Y-MAE \\
\midrule
Modulation bypassed & 20.073 & 0.07365 \\
Uniform weights & 21.588 & 0.08544 \\
Permuted expert weights & 20.574 & 0.09667 \\
Fixed routing prior & 24.463 & 0.05434 \\
Token-hard argmax & 24.367 & 0.05177 \\
Image-mean weights & 24.972 & 0.04919 \\
Spatially shuffled weights & 24.968 & 0.04920 \\
\method & 24.997 & 0.04914 \\
\bottomrule
\end{tabular}
\caption{Modulation and routing interventions on the test set. The permuted row averages seven nonidentity cyclic permutations of the expert-weight assignments.}
\label{tab:mechanism}
\end{table}

\paragraph{Effect of modulation and routing.}
Table~\ref{tab:mechanism} is a separate FP32 run with TF32 disabled; its ordinary forward pass gives 24.997\,dB against the 24.998\,dB of the historical evaluation. Bypassing modulation costs 4.925\,dB, uniform weights 3.409\,dB, and mismatched expert weights 4.423\,dB, while the learned mixture exceeds the fixed prior by 0.534\,dB and lowers Y-MAE from 0.05434 to 0.04914. These interventions establish the role of expert modulation and input-adaptive weighting within this checkpoint.

Spatial averaging and shuffling cost only 0.026 and 0.030\,dB, so image-dependent allocation matters more than where the weights sit. The mixture beats the fixed prior on 58.2\% of individual images, and Fig.~\ref{fig:mechanism_comparison} shows three illustrative scenes.

\begin{figure}[t]
\centering
\includegraphics[width=\textwidth]{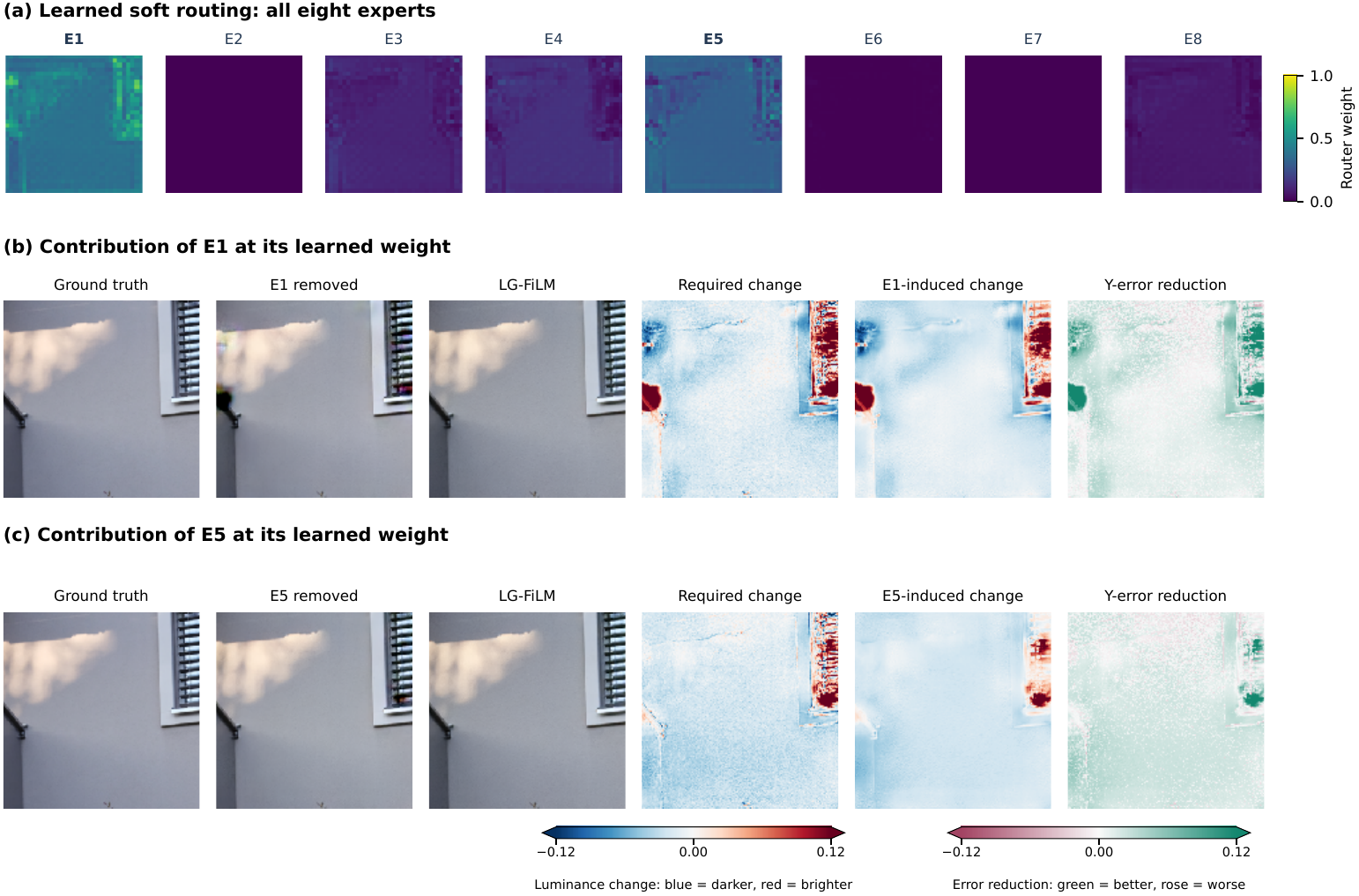}
\caption{Router weights and conditional expert effects for the frozen eight-expert checkpoint on the test set. (a) Eight $32\times32$ weight maps with a common $[0,1]$ scale. (b,c) E1/E5 removal holds all other contributions fixed; required and induced changes are $Y_{\rm GT}-Y_{-k}$ and $Y_{\rm full}-Y_{-k}$ (blue: darker; red: brighter), and error reduction is $|Y_{-k}-Y_{\rm GT}|-|Y_{\rm full}-Y_{\rm GT}|$ (green: better; rose: worse). Output-resolution signed maps use $[-0.12,0.12]$; larger magnitudes saturate.}
\label{fig:mechanism_routing}
\end{figure}

\paragraph{Conditional expert contributions.}
Removing one expert's weighted affine residual while holding all other contributions fixed, without renormalizing, gives
\begin{equation}
F'_{-k}=F'-p_k\bigl(F\odot\gamma_k+\beta_k\bigr).
\end{equation}
Fig.~\ref{fig:mechanism_routing} compares the required change $Y_{\rm GT}-Y_{-k}$ with the induced one $Y_{\rm full}-Y_{-k}$ after the shared decoder and clipping.

On the facade, E1 and E5 hold the largest mean weights (0.404 and 0.297); removing either raises Y-MAE from 0.00980 to 0.02963 or 0.02580 and costs 9.476 or 5.068\,dB RGB PSNR. Over pixels with $|Y_{\rm GT}-Y_{-k}|>0.01$, the induced and required changes share sign on 98.3\% (E1) and 97.9\% (E5), covering 78.9\% and 86.6\% of the image, and absolute luminance error falls on 84.9\% and 85.8\% of all pixels. On the vehicle E1 carries weight 0.713 and its removal costs 10.958\,dB; on fallen leaves E5 carries 0.446 and costs 1.460\,dB. These are illustrative conditional effects, not fixed semantic roles. Every measurement in this subsection keeps the trained weights fixed; Sec.~\ref{sec:sweep_budget} instead compares separately trained variants.

To examine the learned mixture, we compare modulation and routing interventions on the test set. We freeze the trained weights and intervene on the module alone.

\begin{table}[t]
\centering\small
\setlength{\tabcolsep}{5pt}
\begin{tabular}{lrr}
\toprule
Inference intervention & PSNR & $\Delta$ (dB) \\
\midrule
\method & 24.997 & -- \\
Modulation bypassed & 20.073 & $-4.925$ \\
Uniform mixing weights & 21.588 & $-3.409$ \\
Cyclic expert-weight mismatch & 20.574 & $-4.423$ \\
Fixed routing prior & 24.463 & $-0.534$ \\
\bottomrule
\end{tabular}
\caption{Modulation and routing interventions (RGB PSNR, dB). Cyclic mismatch averages seven permutations.}
\label{tab:moe_effect}
\end{table}

\paragraph{Transformation versus gating.}
\label{sec:routing}
With FP32 evaluation and TF32 disabled, \method{} scores 24.997\,dB in Table~\ref{tab:moe_effect}, a 0.001\,dB difference from Table~\ref{tab:sota}. Removing modulation reduces PSNR to 20.073\,dB, a loss of 4.925\,dB. This drop shows that the trained model depends strongly on expert modulation. Replacing learned routing with a fixed prior---the mean routing distribution over the evaluation inputs, computed without labels, with input-conditioned expert transformations left intact---costs only 0.53\,dB. That single comparison separates the two roles the module could be playing. If the model exploited routing as a spatially selective gate, destroying the input dependence of the weights should be catastrophic. This does not happen: the model retains almost all of its quality, because the input dependence it relies on is in the expert outputs, which remain conditioned on $\ctx_i$.

Flattening the weights, by contrast, is expensive: uniform mixing costs 3.41\,dB, more than six times the cost of merely freezing routing. The distinction is that a uniform mixture removes the \emph{relative} emphasis between experts at every token, while a fixed prior preserves it on average; the model needs to know which transformation is currently dominant, not which coordinates it occupies. The same reading explains the last row. Assigning experts the wrong weights---preserving the set of weights at each token but mismatching them to experts through seven cyclic permutations---costs 4.42\,dB, nearly as much as removing the module. The pairing between a weight and the transformation it selects is therefore essential: the experts are not interchangeable, and the model uses their specialization rather than the presence of a mixture alone.

\paragraph{Scene-level adaptation.}
Within the 0.53\,dB that learned routing is worth, the improvement is consistent with scene-level rather than location-level adaptation. Learned routing reduces luminance MAE from 0.05434 to 0.04914 and improves PSNR on 58.2\% of images, that is, on a majority of individual scenes rather than through a few large wins. Spatial averaging and shuffling of the routing weights, which preserve the marginal distribution over experts and destroy only the spatial arrangement, cost 0.026 and 0.030\,dB. How the gate is arranged across the image is nearly irrelevant; that it depends on the image is what matters.

\paragraph{Interpolation versus selection.}
If dense soft fusion is doing real work, hard selection should hurt. Forcing each location to its argmax expert lowers the test PSNR by 0.63\,dB (Table~\ref{tab:oracle}), so the trained model exploits the interior of the simplex rather than a discrete choice. The oracle bounds the other side of the question: a ground-truth-guided choice of the best single expert per image yields 26.040\,dB, 1.04\,dB above the soft mixture, and it is unavailable at deployment. This gap is the cost of committing to one expert per location, and it is small enough that the mixture is not compensating for a weak candidate set.

\begin{table}[t]
\centering\small
\setlength{\tabcolsep}{3pt}
\begin{tabular}{lrrrr}
\toprule
Model & Soft & Token-hard & Oracle & Oracle$-$soft \\
\midrule
\method & 24.998 & 24.367 & 26.040 & $+1.042$ \\
No skips & 24.654 & 23.150 & 24.177 & $-0.477$ \\
Reparameterized & 24.830 & 23.875 & 25.972 & $+1.141$ \\
Decoder cond. & 24.652 & 24.629 & 26.477 & $+1.826$ \\
Cond., $E=16$ & 24.891 & 24.834 & 26.978 & $+2.086$ \\
\bottomrule
\end{tabular}
\caption{Soft fusion, hard selection, and single-expert oracle (RGB PSNR, dB). The last row uses decoder conditioning and sixteen experts. Oracle selections use ground truth.}
\label{tab:oracle}
\end{table}

The same table supplies a useful control. Token-hard selection falls below the soft mixture in every row, and the gap narrows to 0.023\,dB in the decoder-conditioned configuration. Soft fusion is therefore not an artifact of one architecture or of where the mixture is placed; it pays off wherever the mixture is learned.

\paragraph{Are the experts actually specialized?}
A mixture that mixes poorly would still match a single learned transformation if every expert had collapsed onto the same function, and none of the aggregate numbers above would reveal it. Removing one expert at a time settles the question, by deleting $p_k(\feat\odot\gamma_k+\beta_k)$ from the fused feature and leaving every other contribution un-renormalized. If the experts were interchangeable, the loss would be small and roughly equal across indices. It is neither. On a facade, removing E1 or E5, whose mean weights are 0.404 and 0.297, costs 9.48 and 5.07\,dB, and the induced brightness change agrees in sign with the correction the ground truth requires on 98.3\% and 97.9\% of the pixels where that correction exceeds $0.01$. The roles also move with the content: E1 carries weight 0.713 on a vehicle and its removal costs 10.96\,dB, whereas E5 carries 0.446 on fallen leaves at a cost of 1.46\,dB. Experts therefore act as distinct processing modes whose importance is set by the scene, not as eight copies of one affine map. The per-expert maps and their conditional effects appear in Fig.~\ref{fig:mechanism_routing}.

\subsection{Measured Inference Cost}
\label{sec:efficiency}

Parameter counts understate the price of a dense mixture, so we measure. All timings use batch 1, FP32, a single NVIDIA L40S, TF32 disabled, cuDNN benchmarking disabled, 50 warm-up iterations, and 200 timed iterations with tiling disabled.

\begin{table}[t]
\centering\small
\setlength{\tabcolsep}{4pt}
\begin{tabular}{lrrr}
\toprule
Configuration & Resolution & Latency (ms) & FPS \\
\midrule
Full, $E{=}8$ & $256\times256$ & 3.15 & 318 \\
Full, $E{=}8$ & $1920\times1080$ & 17.23 & 58.0 \\
Full, $E{=}8$ & $3968\times2976$ & 131.11 & 7.6 \\
Modulation bypassed & $256\times256$ & 1.88 & 532 \\
\bottomrule
\end{tabular}
\caption{Measured forward latency. At $256\times256$ the full model has median 2.96\,ms and $p95$ 4.08\,ms. Peak allocated memory: 35.7\,MB, 626.5\,MB, and 4{,}190.9\,MB at the three resolutions.}
\label{tab:latency}
\end{table}

Table~\ref{tab:latency} reports the full model at three resolutions. At $256\times256$ a forward pass takes 3.15\,ms on average with a median of 2.96\,ms and a $p95$ of 4.08\,ms, at 35.7\,MB peak memory; at $1920\times1080$ it takes 17.23\,ms at 626.5\,MB; and at the ZRR native resolution of $3968\times2976$ it takes 131.11\,ms at 4{,}190.9\,MB. Bypassing modulation reduces the $256\times256$ pass to 1.88\,ms, so the conditional branch costs 1.27\,ms---roughly 40\% of end-to-end time for 16\% of the parameters. Within one pass the encoder takes 0.888\,ms, the bottleneck stage including the module 1.391\,ms, and the decoder 0.708\,ms. Full-resolution inference fits in a single pass without tiling, and the $6.7\times$ memory growth from $1920\times1080$ to $3968\times2976$ exceeds the $5.7\times$ pixel-count ratio because full-resolution feature maps, not the module, dominate it.

The compute cost shows the same pattern. Counting only convolution, matrix--matrix and matrix--vector products, the full model needs 5.763\,GFLOPs at $256\times256$ and 1{,}038.5\,GFLOPs at $3968\times2976$, of which the conditional bottleneck accounts for 1.215\,G (21.1\%) and 218.8\,G (21.1\%) respectively---the share is invariant to resolution because both the module and the backbone scale with token count. Doubling the expert bank to sixteen raises the module's cost to 1.756\,G at $256\times256$, an increase of 44.5\%, while the encoder and decoder are unchanged, confirming that expert count enters linearly and only through this branch. The module accounts for 21.1\% of the arithmetic but 44.2\% of the wall-clock time, because it is a stack of many small per-token matrix products rather than a few large convolutions and therefore reaches a smaller fraction of peak throughput. A natural next step is fewer participating tokens or a narrower expert.

\begin{table}[t]
\centering\small
\setlength{\tabcolsep}{4pt}
\begin{tabular}{lrrr}
\toprule
Module & \multicolumn{2}{c}{$E{=}8$} & \multicolumn{1}{c}{$E{=}16$} \\
\cmidrule(lr){2-3}
 & $256^2$ & $3968{\times}2976$ & $256^2$ \\
\midrule
Encoder & 2.624 (45.5\%) & 472.7 (45.5\%) & 2.624 \\
Cond.\ bottleneck & 1.215 (21.1\%) & 218.8 (21.1\%) & 1.756 \\
Decoder & 1.812 (31.4\%) & 326.5 (31.4\%) & 1.812 \\
Head & 0.113 (2.0\%) & 20.4 (2.0\%) & 0.113 \\
\midrule
Total & 5.763 & 1038.5 & 6.304 \\
\bottomrule
\end{tabular}
\caption{Measured multiply--accumulates (GFLOPs) by module. Counts cover convolutions and matrix products only; elementwise operations, activations, softmax, and the affine fusion are excluded, so the module's true cost is under-stated. Percentages refer to the $E{=}8$ columns.}
\label{tab:flops}
\end{table}

\subsection{Qualitative Results}
\label{sec:qualitative}

Qualitative comparisons appear in Figs.~\ref{fig:qualitative}--\ref{fig:coco}. On six Fujifilm UltraISP scenes spanning sky, vegetation, a facade, a vehicle, graffiti, and a painted container, the model tracks brightness and color rendition while retaining local structure. Fig.~\ref{fig:mechanism_comparison} contrasts the mixture with bypassed modulation, which distorts wall tones and adds artifacts, and with the fixed prior, which recovers global appearance but leaves larger local errors.

\section{Discussion and Conclusion}
\label{sec:limits}

\paragraph{Why the adaptation is scene-level.}
Taken together, the interventions describe a module that learns a small vocabulary of processing modes and selects among them per scene, rather than a module that assigns different treatments to different image regions. This is the shape one should expect. Illumination and color rendition are properties of the scene: the sky, the facade, and the foliage in one photograph share them, and an ISP that corrected them per location would estimate a quantity that is nearly constant wherever it is not degenerate. Local evidence has a different job---it modulates \emph{how strongly} the chosen mode is applied and where detail must be preserved. Our design allocates the two accordingly: the global descriptor sets the mode, the local term perturbs it. A gate that selected experts mainly by location would be solving a segmentation problem that the reconstruction objective never asks for and supplies no supervision for. The measurements agree, and they explain why uniform weights hurt far more than a fixed prior: averaging destroys the mode, whereas fixing the prior forfeits only its per-scene adjustment.

\paragraph{Limitations.}
Expert count and conditioning change together in the expanded configuration, and the differences in Table~\ref{tab:architecture_summary} are single runs. Dense evaluation of all experts keeps cost linear in $E$ and, as measured, substantial at high resolution. The COCO-derived benchmark is synthesized, so it probes pipeline behavior rather than real-sensor generalization. Fujifilm UltraISP mainly contains daytime scenes. Evaluation under more diverse lighting conditions would further assess scene-adaptive processing.

\paragraph{Conclusion.}
We formulated input-adaptive ISP processing as a dense mixture of affine transformations acting on a shared bottleneck, conditioned jointly on local neighborhood and global scene evidence. The design makes adaptation an explicit object of the network: it can be bypassed, its routing replaced or permuted, and its experts removed individually, each intervention producing a measurable and interpretable change. Those interventions locate the model's advantage in learned transformations that adapt per scene rather than in spatially selective gating, while the shared decoder and skips carry reconstruction. Reporting measured latency alongside the parameter count makes the trade explicit: the mixture is compact in parameters and substantial in compute, a balance a practical image signal processor must evaluate.

\label{sec:main_end}

{\small
\bibliographystyle{unsrt}
\bibliography{references}

\begin{thebibliography}{10}

\bibitem{reinhard2002photographic}
Erik Reinhard, Michael Stark, Peter Shirley, and James Ferwerda.
\newblock Photographic tone reproduction for digital images.
\newblock {\em ACM Transactions on Graphics}, 21(3):267--276, 2002.

\bibitem{durand2002bilateral}
Fr{\'e}do Durand and Julie Dorsey.
\newblock Fast bilateral filtering for the display of high-dynamic-range images.
\newblock {\em ACM Transactions on Graphics}, 21(3):257--266, 2002.

\bibitem{schwartz2019deepisp}
Eli Schwartz, Raja Giryes, and Alex~M. Bronstein.
\newblock {DeepISP}: Towards learning an end-to-end image processing pipeline.
\newblock {\em IEEE Transactions on Image Processing}, 28(2):912--923, 2019.

\bibitem{chen2018seeing}
Chen Chen, Qifeng Chen, Jia Xu, and Vladlen Koltun.
\newblock Learning to see in the dark.
\newblock In {\em Proceedings of the IEEE Conference on Computer Vision and Pattern Recognition}, 2018.

\bibitem{ignatov2020pynet}
Andrey Ignatov, Luc~Van Gool, and Radu Timofte.
\newblock Replacing mobile camera {ISP} with a single deep learning model.
\newblock In {\em Proceedings of the IEEE/CVF Conference on Computer Vision and Pattern Recognition Workshops}, 2020.

\bibitem{li2024rmfa}
Fei Li, Wenbo Hou, and Peng Jia.
\newblock {RMFA-Net}: A neural {ISP} for real {RAW} to {RGB} image reconstruction.
\newblock {\em arXiv preprint arXiv:2406.11469}, 2024.

\bibitem{souza2023metaisp}
Matheus Souza and Wolfgang Heidrich.
\newblock {MetaISP}---exploiting global scene structure for accurate multi-device color rendition.
\newblock In {\em Vision, Modeling, and Visualization}. The Eurographics Association, 2023.

\bibitem{ignatov2022pynetv2}
Andrey Ignatov, Grigory Malivenko, Radu Timofte, Yu~Tseng, Yu-Syuan Xu, Po-Hsiang Yu, Cheng-Ming Chiang, Hsien-Kai Kuo, Min-Hung Chen, Chia-Ming Cheng, and Luc~Van Gool.
\newblock {PyNET-V2 Mobile}: Efficient on-device photo processing with neural networks.
\newblock {\em arXiv preprint arXiv:2211.06263}, 2022.

\bibitem{kim2020pynetca}
Byung-Hoon Kim, Joonyoung Song, Jong~Chul Ye, and JaeHyun Baek.
\newblock {PyNET-CA}: Enhanced {PyNET} with channel attention for end-to-end mobile image signal processing.
\newblock In {\em Computer Vision -- ECCV 2020 Workshops}, 2020.

\bibitem{dai2020awnet}
Linhui Dai, Xiaohong Liu, Chengqi Li, and Jun Chen.
\newblock {AWNet}: Attentive wavelet network for image {ISP}.
\newblock In {\em Computer Vision -- ECCV 2020 Workshops}, 2020.

\bibitem{zhang2021aligned}
Zhilu Zhang, Haolin Wang, Ming Liu, Ruohao Wang, Jiawei Zhang, and Wangmeng Zuo.
\newblock Learning {RAW}-to-{sRGB} mappings with inaccurately aligned supervision.
\newblock In {\em Proceedings of the IEEE/CVF International Conference on Computer Vision}, 2021.

\bibitem{ignatov2022microisp}
Andrey Ignatov, Anastasia Sycheva, Radu Timofte, Yu~Tseng, Yu-Syuan Xu, Po-Hsiang Yu, Cheng-Ming Chiang, Hsien-Kai Kuo, Min-Hung Chen, Chia-Ming Cheng, and Luc~Van Gool.
\newblock {MicroISP}: Processing {32MP} photos on mobile devices with deep learning.
\newblock {\em arXiv preprint arXiv:2211.06770}, 2022.

\bibitem{ignatov2025mobileisp}
Andrey Ignatov, Georgy Perevozchikov, Radu Timofte, et~al.
\newblock Learned smartphone {ISP} on mobile {GPUs}, mobile {AI} 2025 challenge: Report.
\newblock In {\em Proceedings of the IEEE/CVF Conference on Computer Vision and Pattern Recognition Workshops}, pages 1934--1946, 2025.

\bibitem{ronneberger2015unet}
Olaf Ronneberger, Philipp Fischer, and Thomas Brox.
\newblock {U-Net}: Convolutional networks for biomedical image segmentation.
\newblock In {\em Medical Image Computing and Computer-Assisted Intervention}, pages 234--241, 2015.

\bibitem{chen2022simple}
Liangyu Chen, Xiaojie Chu, Xiangyu Zhang, and Jian Sun.
\newblock Simple baselines for image restoration.
\newblock In {\em European Conference on Computer Vision}, pages 17--33, 2022.

\bibitem{ding2021repvgg}
Xiaohan Ding, Xiangyu Zhang, Ningning Ma, Jungong Han, Guiguang Ding, and Jian Sun.
\newblock {RepVGG}: Making {VGG}-style {ConvNets} great again.
\newblock In {\em Proceedings of the IEEE/CVF Conference on Computer Vision and Pattern Recognition}, pages 13733--13742, 2021.

\bibitem{brooks2019unprocessing}
Tim Brooks, Ben Mildenhall, Tianfan Xue, Jiawen Chen, Dillon Sharlet, and Jonathan~T. Barron.
\newblock Unprocessing images for learned raw denoising.
\newblock In {\em Proceedings of the IEEE/CVF Conference on Computer Vision and Pattern Recognition}, 2019.

\bibitem{xing2021invertible}
Yazhou Xing, Zian Qian, and Qifeng Chen.
\newblock Invertible image signal processing.
\newblock In {\em Proceedings of the IEEE/CVF Conference on Computer Vision and Pattern Recognition}, 2021.

\bibitem{perez2018film}
Ethan Perez, Florian Strub, Harm de~Vries, Vincent Dumoulin, and Aaron Courville.
\newblock {FiLM}: Visual reasoning with a general conditioning layer.
\newblock In {\em Proceedings of the AAAI Conference on Artificial Intelligence}, 2018.

\bibitem{huang2017adain}
Xun Huang and Serge Belongie.
\newblock Arbitrary style transfer in real-time with adaptive instance normalization.
\newblock In {\em Proceedings of the IEEE International Conference on Computer Vision}, 2017.

\bibitem{park2019spade}
Taesung Park, Ming-Yu Liu, Ting-Chun Wang, and Jun-Yan Zhu.
\newblock Semantic image synthesis with spatially-adaptive normalization.
\newblock In {\em Proceedings of the IEEE/CVF Conference on Computer Vision and Pattern Recognition}, pages 2337--2346, 2019.

\bibitem{hu2018squeeze}
Jie Hu, Li~Shen, and Gang Sun.
\newblock Squeeze-and-excitation networks.
\newblock In {\em Proceedings of the IEEE Conference on Computer Vision and Pattern Recognition}, 2018.

\bibitem{gharbi2017bilateral}
Micha{\"e}l Gharbi, Jiawen Chen, Jonathan~T. Barron, Samuel~W. Hasinoff, and Fr{\'e}do Durand.
\newblock Deep bilateral learning for real-time image enhancement.
\newblock {\em ACM Transactions on Graphics}, 36(4), 2017.

\bibitem{jacobs1991adaptive}
Robert~A. Jacobs, Michael~I. Jordan, Steven~J. Nowlan, and Geoffrey~E. Hinton.
\newblock Adaptive mixtures of local experts.
\newblock {\em Neural Computation}, 3(1):79--87, 1991.

\bibitem{shazeer2017outrageously}
Noam Shazeer, Azalia Mirhoseini, Krzysztof Maziarz, Andy Davis, Quoc Le, Geoffrey Hinton, and Jeff Dean.
\newblock Outrageously large neural networks: The sparsely-gated mixture-of-experts layer.
\newblock In {\em International Conference on Learning Representations}, 2017.

\bibitem{fedus2022switch}
William Fedus, Barret Zoph, and Noam Shazeer.
\newblock {Switch Transformers}: Scaling to trillion parameter models with simple and efficient sparsity.
\newblock {\em Journal of Machine Learning Research}, 23(120):1--39, 2022.

\bibitem{riquelme2021vision}
Carlos Riquelme, Joan Puigcerver, Basil Mustafa, Maxim Neumann, Rodolphe Jenatton, Andr{\'e}~Susano Pinto, Daniel Keysers, and Neil Houlsby.
\newblock Scaling vision with sparse mixture of experts.
\newblock In {\em Advances in Neural Information Processing Systems}, 2021.

\bibitem{zhou2022expert}
Yanqi Zhou, Tao Lei, Hanxiao Liu, Nan Du, Yanping Huang, Vincent Zhao, Andrew Dai, Zhifeng Chen, Quoc Le, and James Laudon.
\newblock Mixture-of-experts with expert choice routing.
\newblock In {\em Advances in Neural Information Processing Systems}, 2022.

\bibitem{puigcerver2024soft}
Joan Puigcerver, Carlos Riquelme, Basil Mustafa, and Neil Houlsby.
\newblock From sparse to soft mixtures of experts.
\newblock In {\em International Conference on Learning Representations}, 2024.

\bibitem{yang2019condconv}
Brandon Yang, Gabriel Bender, Quoc~V. Le, and Jiquan Ngiam.
\newblock {CondConv}: Conditionally parameterized convolutions for efficient inference.
\newblock In {\em Advances in Neural Information Processing Systems}, 2019.

\bibitem{chen2020dynamic}
Yinpeng Chen, Xiyang Dai, Mengchen Liu, Dongdong Chen, Lu~Yuan, and Zicheng Liu.
\newblock Dynamic convolution: Attention over convolution kernels.
\newblock In {\em Proceedings of the IEEE/CVF Conference on Computer Vision and Pattern Recognition}, 2020.

\bibitem{zamfir2025moce}
Eduard Zamfir, Zongwei Wu, Nancy Mehta, Yuedong Tan, Danda~Pani Paudel, Yulun Zhang, and Radu Timofte.
\newblock Complexity experts are task-discriminative learners for any image restoration.
\newblock In {\em Proceedings of the IEEE/CVF Conference on Computer Vision and Pattern Recognition}, 2025.

\bibitem{lin2014coco}
Tsung-Yi Lin, Michael Maire, Serge Belongie, James Hays, Pietro Perona, Deva Ramanan, Piotr Doll{\'a}r, and C.~Lawrence Zitnick.
\newblock {Microsoft COCO}: Common objects in context.
\newblock In {\em European Conference on Computer Vision}, pages 740--755, 2014.

\bibitem{wang2004ssim}
Zhou Wang, Alan~C. Bovik, Hamid~R. Sheikh, and Eero~P. Simoncelli.
\newblock Image quality assessment: From error visibility to structural similarity.
\newblock {\em IEEE Transactions on Image Processing}, 13(4):600--612, 2004.

\bibitem{zhang2018lpips}
Richard Zhang, Phillip Isola, Alexei~A. Efros, Eli Shechtman, and Oliver Wang.
\newblock The unreasonable effectiveness of deep features as a perceptual metric.
\newblock In {\em Proceedings of the IEEE Conference on Computer Vision and Pattern Recognition}, pages 586--595, 2018.

\bibitem{mai2026mobileisp}
Andrey Ignatov, Georgy Perevozchikov, Radu Timofte, Yuhong He, Yuxuan Qiu, Tianyu Hao, Ruoqi Li, Yueqi Yang, Chaoyu Feng, et~al.
\newblock Learned smartphone {ISP} on mobile {GPUs}: Mobile {AI} 2026 challenge report.
\newblock In {\em arXiv preprint arXiv:2611.05008}, 2026.

\end{thebibliography}
}
\end{document}